\documentclass[journal]{IEEEtran}
\usepackage{graphicx}
\usepackage{multirow}
\usepackage{adjustbox}
\usepackage{amsfonts}
\usepackage{xcolor}
\usepackage{subcaption}
\usepackage{hyperref}
\usepackage{amsmath}
\usepackage{cite}

\begin{document}
\title{SBERT-WK: A Sentence Embedding Method by Dissecting BERT-based Word Models}

\author{
Bin~Wang,~\IEEEmembership{Student Member,~IEEE,}
and~C.-C.~Jay~Kuo,~\IEEEmembership{Fellow,~IEEE}
\thanks{Bin Wang is with Ming-Hsieh Department of Electrical and
Computer Engineering and Signal and Image Processing Institute,
University of Southern California, Los Angeles, CA, 90089-2584 USA
(e-mail:bwang28c@gmail.com).}
\thanks{C.-C. Jay Kuo is with Ming-Hsieh Department of Electrical and
Computer Engineering, Department of Computer Science and Signal and
Image Processing Institute, University of Southern California, Los
Angeles, CA, 90089-2584 USA (e-mail:cckuo@sipi.usc.edu).} }

\markboth{Journal of \LaTeX\ Class Files,~Vol.~14, No.~8, August~2020}%
{Shell \MakeLowercase{\textit{et al.}}: Bare Demo of IEEEtran.cls for IEEE Journals}

\maketitle

\begin{abstract}

Sentence embedding is an important research topic in natural language
processing (NLP) since it can transfer knowledge to downstream tasks.
Meanwhile, a contextualized word representation, called BERT, achieves
the state-of-the-art performance in quite a few NLP tasks. Yet, it is an
open problem to generate a high quality sentence representation from
BERT-based word models. It was shown in previous study that different
layers of BERT capture different linguistic properties. This allows us
to fuse information across layers to find better sentence
representations. In this work, we study the layer-wise pattern of the
word representation of deep contextualized models. Then, we propose a
new sentence embedding method by dissecting BERT-based word models
through geometric analysis of the space spanned by the word
representation. It is called the SBERT-WK method\footnote{Named after the author's last name initials.}. No further training is
required in SBERT-WK.  We evaluate SBERT-WK on semantic textual
similarity and downstream supervised tasks. Furthermore, ten
sentence-level probing tasks are presented for detailed linguistic
analysis. Experiments show that SBERT-WK achieves the state-of-the-art
performance. Our codes are publicly available\footnote{
\href{https://github.com/BinWang28/BERT_Sentence_Embedding}{https://github.com/BinWang28/SBERT-WK-Sentence-Embedding}.}. 

\end{abstract}


\IEEEpeerreviewmaketitle

\section{Introduction}\label{sec:introduction}

\IEEEPARstart{S}{tatic} word embedding is a popular learning technique
that transfers prior knowledge from a large unlabeled corpus
\cite{word2vec, GloVe, Fasttext}. Most of recent sentence embedding
methods are rooted in that static word representations can be embedded
with rich syntactic and semantic information. It is desired to extend
the word-level embedding to the sentence-level, which contains a longer
piece of text.  We have witnessed a breakthrough by replacing the
``static" word embedding to the ``contextualized" word embedding in
the last several years, e.g., \cite{ELMO, BERT,XLNET,GPT}.  A natural
question to ask is how to exploit contextualized word embedding in the
context of sentence embedding. Here, we examine the problem of learning
the universal representation of sentences. A contextualized word
representation, called BERT, achieves the state-of-the-art performance
in many natural language processing (NLP) tasks. We aim to develop a
sentence embedding solution from BERT-based models in this work. 

As reported in \cite{BERT_STRU} and \cite{liu2019linguistic}, different
layers of BERT learn different levels of information and linguistic
properties. While intermediate layers encode the most transferable
features, representation from higher layers are more expressive in
high-level semantic information.  Thus, information fusion across layers
has its potential to provide a stronger representation. Furthermore,
by conducting experiments on patterns of the isolated word
representation across layers in deep models, we observe the following
property. Words of richer information in a sentence have higher
variation in their representations, while the token representation
changes gradually, across layers. This finding helps define ``salient"
word representations and informative words in computing universal
sentence embedding. 

One limitation of BERT is that due to the large model size, it is time consuming to perform
sentence pair regression such as clustering and semantic search. One effective way to solve this problem is to transforms a sentence to a vector that encodes
the semantic meaning of the sentence. Currently, a common sentence
embedding approach from BERT-based models is to average the
representations obtained from the last layer or using the [CLS]
token for sentence-level prediction. Yet, both are sub-optimal as shown
in the experimental section of this paper. To the best of our knowledge,
there is only one paper on sentence embedding using pre-trained BERT,
called SBERT \cite{reimers2019sentence}. It leverages
further training with high-quality labeled sentence pairs.  Apparently,
how to obtain sentence embedding from deep contextualized models is
still an open problem. 


Different from SBERT, we
investigate sentence embedding by studying the geometric structure of
deep contextualized models and propose a new method by dissecting
BERT-based word models. It is called the SBERT-WK method. SBERT-WK inherits the strength of deep contextualized models which is trained on both word- and sentence-level objectives. It is
compatible with most deep contextualized models such as BERT \cite{BERT}
and RoBERTa \cite{liu2019roberta}. 

This work has the following three main contributions.
\begin{enumerate}
\item We study the evolution of isolated word representation patterns
across layers in BERT-based models.  These patterns are shown to be
highly correlated with word's content. It provides useful insights
into deep contextualized word models. 
\item We propose a new sentence embedding method, called SBERT-WK,
through geometric analysis of the space learned by deep contextualized
models. 
\item We evaluate the SBERT-WK method against eight downstream tasks and
seven semantic textual similarity tasks, and show that it achieves
state-of-the-art performance. Furthermore, we use sentence-level probing
tasks to shed light on the linguistic properties learned by SBERT-WK. 
\end{enumerate}

The rest of the paper is organized as following. Related work is
reviewed in Sec. \ref{sec:related_work}. The evolution of word
representation patterns in deep contextualized models is studied in Sec.
\ref{sec:patten}. The proposed SBERT-WK method is presented in Sec.
\ref{sec:dissecting_bert}. The SBERT-WK method is evaluated with respect
to various tasks in Sec. \ref{sec:experiments}. Finally, concluding
remarks and future work directions are given in Sec.
\ref{sec:conclusion}. 

\section{Related Work} \label{sec:related_work}

\subsection{Contextualized Word Embedding}

Traditional word embedding methods provide a static representation for a
word in a vocabulary set.  Although the static representation is widely
adopted in NLP, it has several limitations in modeling the context
information. First, it cannot deal with polysemy. Second, it cannot
adjust the meaning of a word based on its contexts. To address the
shortcomings of static word embedding methods, there is a new trend to
go from shallow to deep contextualized representations.  For example,
ELMo \cite{ELMO}, GPT \cite{GPT} and BERT \cite{BERT}
are pre-trained deep neural language models, and they can be fine-tuned
on specific tasks. These new word embedding methods achieve impressive
performance on a wide range of NLP tasks. In particular, the BERT-based
models are dominating in leaderboards of language understanding tasks
such as SQuAD2.0 \cite{SQUAD_2} and GLUE benchmarks \cite{wang2018glue}. 

ELMo is one of the earlier work in applying a pre-trained language model
to downstream tasks \cite{ELMO}. It employs two layer bi-directional
LSTM and fuses features from all LSTM outputs using task-specific
weights. OpenAI GPT \cite{GPT} incorporates a fine-tuning process when
it is applied to downstream tasks. Task-specific parameters are
introduced and fine-tuned with all pre-trained parameters. BERT employs
the Transformer architecture \cite{transformer}, which is composed by
multiple multi-head attention layers. It can be trained more efficiently
than LSTM. It is trained on a large unlabeled corpus with several
objectives to learn both word- and sentence-level information, where the
objectives include masked language modeling as well as the next sentence
prediction. A couple of variants have been proposed based on BERT.
RoBERTa \cite{liu2019roberta} attempts to improve BERT by providing a
better recipe in BERT model training.  ALBERT \cite{lan2019albert}
targets at compressing the model size of BERT by introducing two
parameter-reduction techniques. At the same time, it achieves better
performance.  XLNET \cite{XLNET} adopts a generalized auto-regressive
pre-training method that has the merits of auto-regressive and
auto-encoder language models. 

Because of the superior performance of BERT-based models, it is
important to have a better understanding of BERT-based models and the
transformer architecture. Efforts have been made along this direction
recently as reviewed below.  Liu {\em et al.} \cite{liu2019linguistic}
and Petroni {\em et al.} \cite{petroni2019language} used word-level
probing tasks to investigate the linguistic properties learned by the
contextualized models experimentally.  Kovaleva {\em et al.}
\cite{kovaleva2019revealing} and Michel {\em et al.}
\cite{michel2019sixteen} attempted to understand the self-attention
scheme in BERT-based models. Hao {\em et al.} \cite{hao2019visualizing}
provided insights into BERT by visualizing and analyzing the loss
landscapes in the fine-tuning process. Ethayarajh
\cite{ethayarajh2019contextual} explained how the deep contextualized
model learns the context representation of words. Despite the
above-mentioned efforts, the evolving pattern of a word representation
across layers in BERT-based models has not been studied before. In this
work, we first examine the pattern evolution of a token representation
across layers without taking its context into account.  With the
context-independent analysis, we observe that the evolving patterns are
highly related to word properties. This observation in turn inspires the
proposal of a new sentence embedding method -- SBERT-WK. 

\subsection{Universal Sentence Embedding}

By sentence embedding, we aim at extracting a numerical representation
for a sentence to encapsulate its meanings. The linguistic features
learned by a sentence embedding method can be external information
resources for downstream tasks. Sentence embedding methods can be
categorized into two categories: non-parameterized and parameterized
models.  Non-parameterized methods usually rely on high quality
pre-trained word embedding methods. Following
this line of averaging word embeddings, several weighted averaging methods were proposed,
including tf-idf, SIF \cite{arora2016simple}, uSIF
\cite{ethayarajh2018unsupervised} and GEM \cite{yang2019parameter}. SIF
uses the random walk to model the sentence generation process and
derives word weights using the maximum likelihood estimation (MLE). By exploiting geometric
analysis of the space spanned by word embeddings, GEM determines word
weights with several hand-crafted measurements. 

Parameterized models are more complex, and they usualy perform better
than non-parameterized models. The skip-thought model
\cite{kiros2015skip} extends the unsupervised training of word2vec
\cite{word2vec} from the word level to the sentence level. InferSent
\cite{conneau2017supervised} employs bi-directional LSTM with supervised
training. It trains the model to predict the entailment or contradiction
of sentence pairs with the Stanford Natural Language Inference (SNLI)
dataset. It achieves better results than methods with unsupervised
learning. The SBERT method \cite{reimers2019sentence} is the only parameterized
sentence embedding model using BERT as the backbone. SBERT shares high
similarity with InferSent \cite{conneau2017supervised}. It uses the
Siamese network on top of the BERT model and fine-tunes it based on high
quality sentence inference data (e.g. the SNLI dataset) to learn more
sentence-level information.  However, unlike supervised tasks, universal
sentence embedding methods in general do not have a clear objective
function to optimize. Instead of training on more sophisticated
multi-tasking objectives, we combine the advantage of both parameterized
and non-parameterized methods.  SBERT-WK is computed by subspace
analysis of the manifold learned by the parameterized BERT-based models.

GRAN \cite{wieting2017revisiting} introduced a novel Gated Recurrent Averaging Network that combines benefit of the simple averaging and LSTM and achieved good result on STS tasks. Other than SNLI dataset, GRAN also incorporates both phrase pairs and sentence pairs as the supervision. In contrast, we only use sentence-level objective for fine-tuning BERT-based models and would leave the usage of phrase-level information as future work.

Subspace analysis has already been applied to sentence
embedding in non-parameterized models, GEM is the most related work with ours. Both methods incorporates Gram-Schmidt process in analyzing the word embedding space in order to determine the weights. GEM is built upon static word embedding methods. In contrast, SBERT-WK focuses on more sophisticated deep contextualized representations. Comparisons among both models are also discussed in experimental section.

To the best of our knowledge, our work is the first one that exploits subspace analysis to find generic sentence embedding based on deep contextualized models. We will show in this work
that SBERT-WK can consistently outperform state-of-the-art methods with
low computational overhead and good interpretability, which is
attributed to high transparency and efficiency of subspace analysis and
the power of deep contextualized word embedding. 

\begin{figure*}[htb]
    \centering
    \begin{subfigure}{.3\textwidth}
        \centering
        \includegraphics[width=\textwidth]{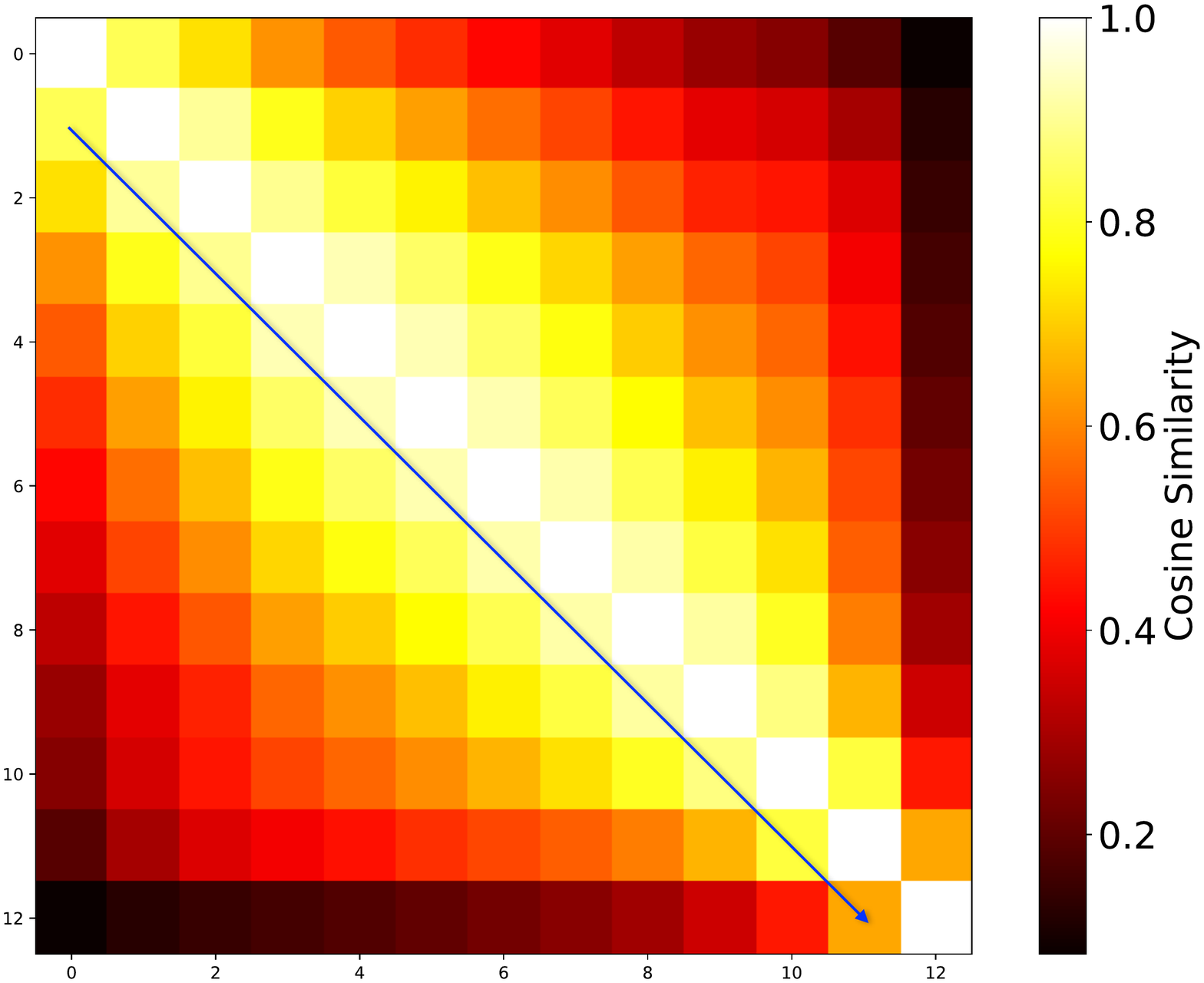}
        \caption{BERT}
        \label{fig:q1_1}
    \end{subfigure}
    \begin{subfigure}{.3\textwidth}
        \centering
        \includegraphics[width=\textwidth]{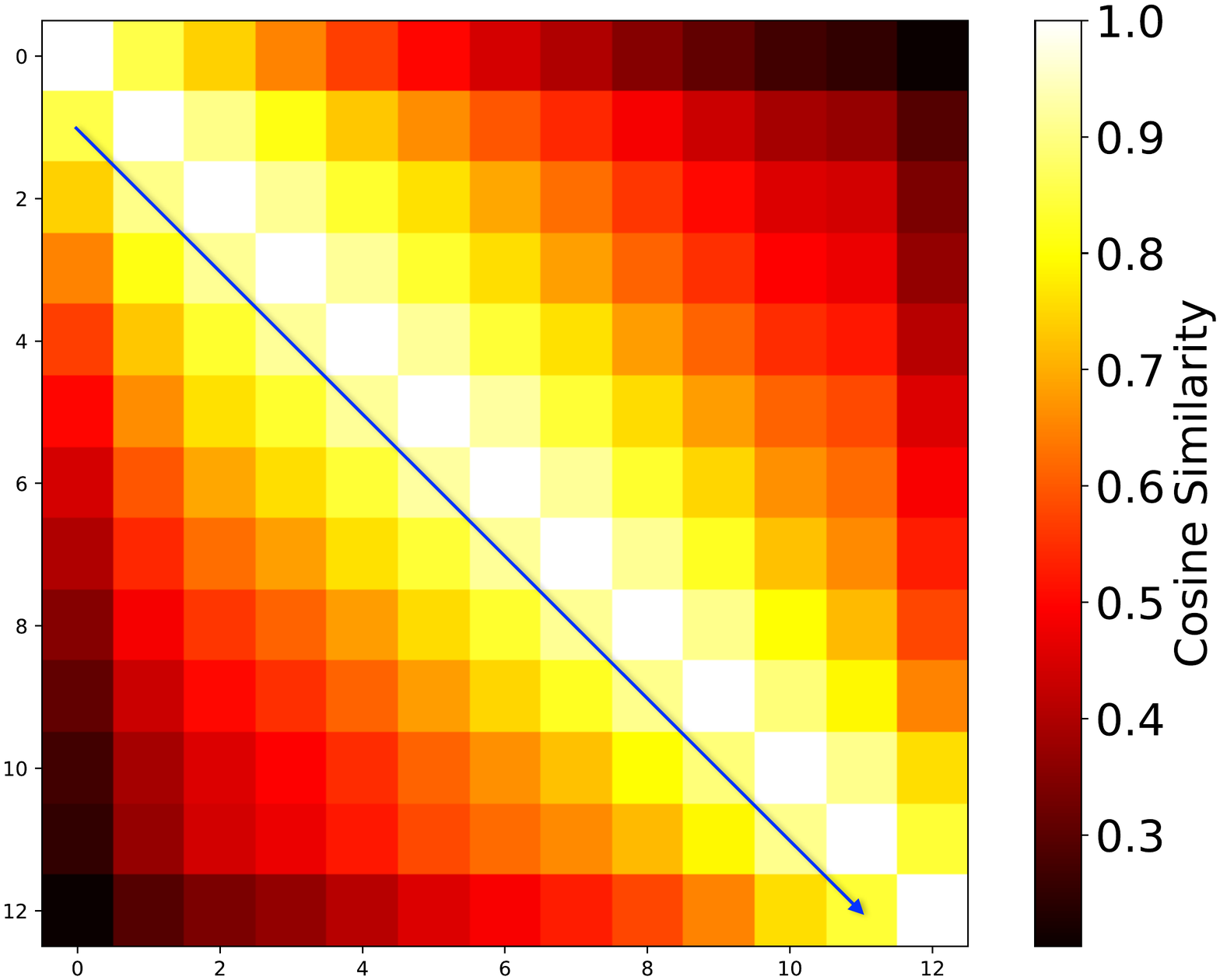}
        \caption{SBERT}
        \label{fig:q1_2}
    \end{subfigure}
    \begin{subfigure}{.3\textwidth}
        \centering
        \includegraphics[width=\textwidth]{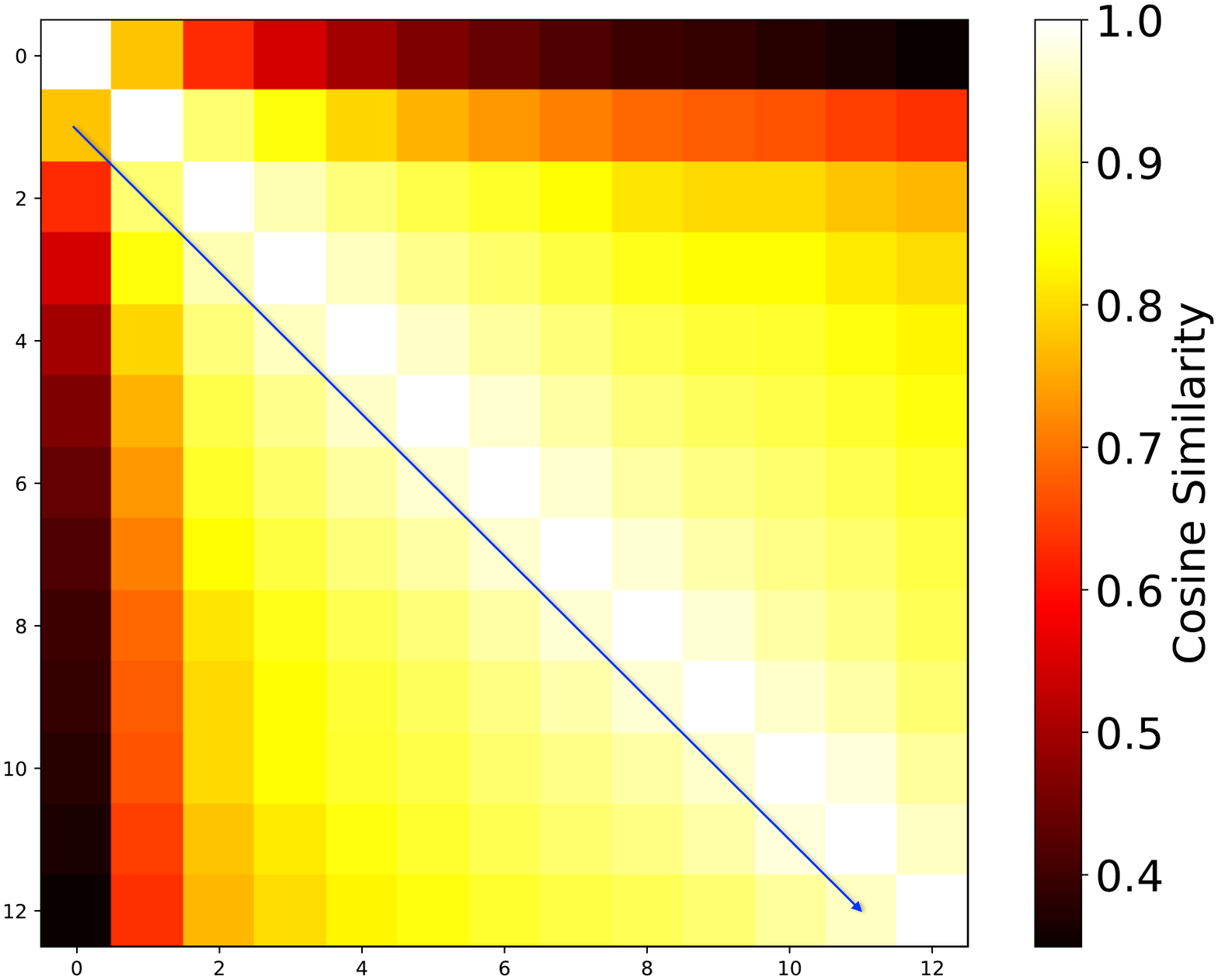}
        \caption{RoBERTa}
        \label{fig:q1_3}
    \end{subfigure}
    \begin{subfigure}{.3\textwidth}
        \centering
        \includegraphics[width=\textwidth]{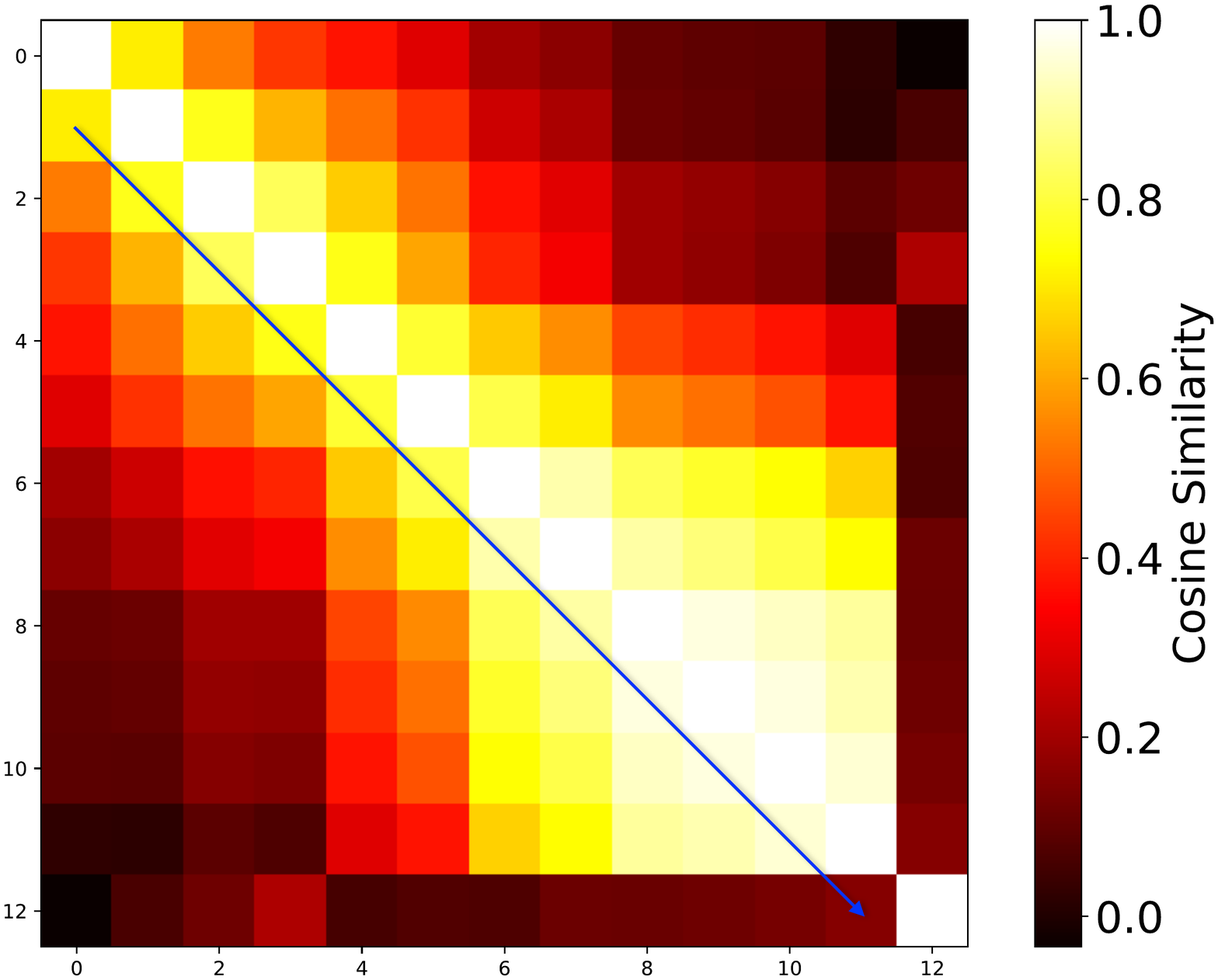}
        \caption{XLNET}
        \label{fig:q1_4}
    \end{subfigure}
    \begin{subfigure}{.3\textwidth}
        \centering
        \includegraphics[width=\textwidth]{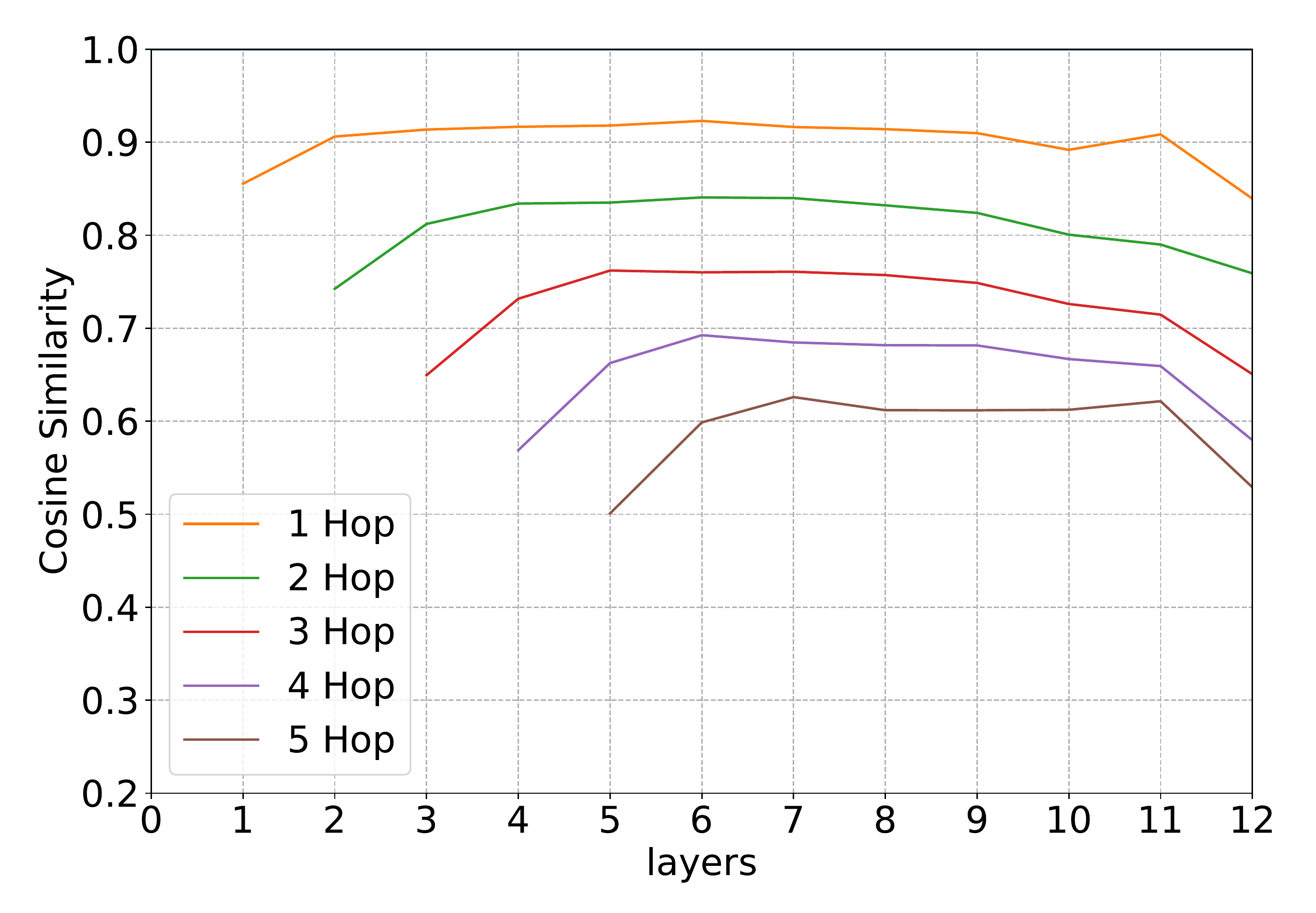}
        \caption{BERT}
        \label{fig:q1_5}
    \end{subfigure}
    \begin{subfigure}{.3\textwidth}
        \centering
        \includegraphics[width=\textwidth]{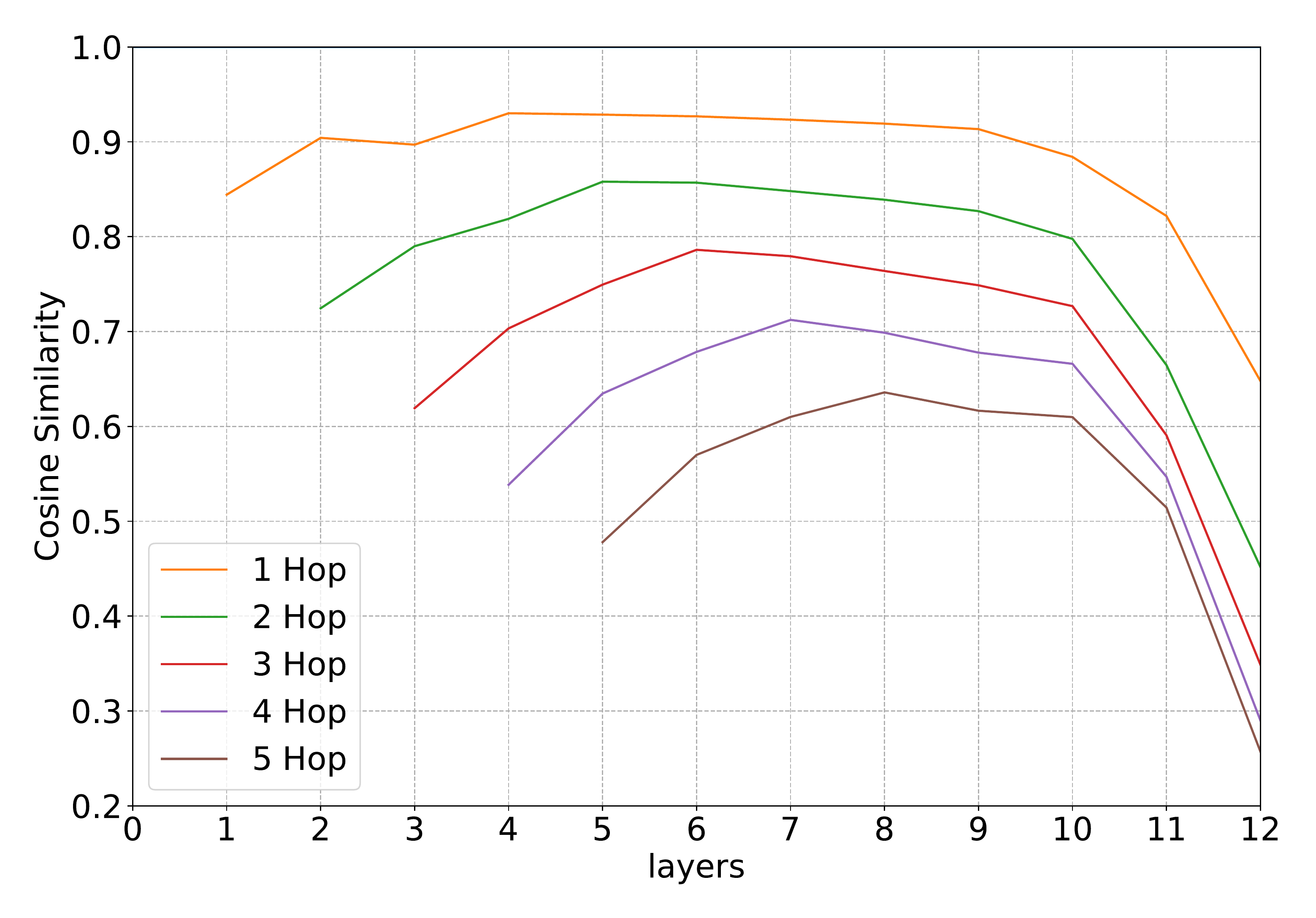}
        \caption{SBERT}
        \label{fig:q1_6}
    \end{subfigure}
    \begin{subfigure}{.3\textwidth}
        \centering
        \includegraphics[width=\textwidth]{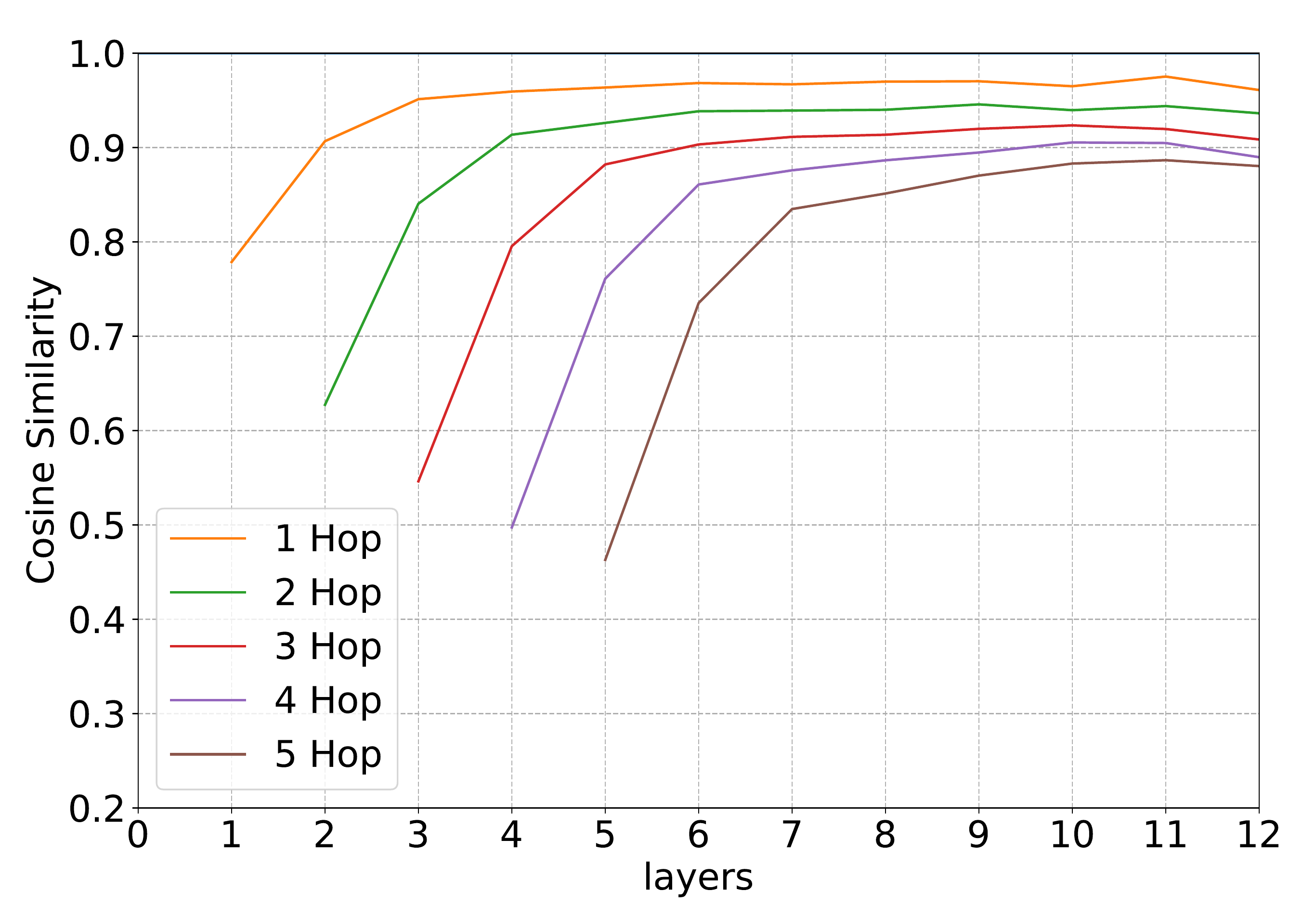}
        \caption{RoBERTa}
        \label{fig:q1_7}
    \end{subfigure}
    \begin{subfigure}{.3\textwidth}
        \centering
        \includegraphics[width=\textwidth]{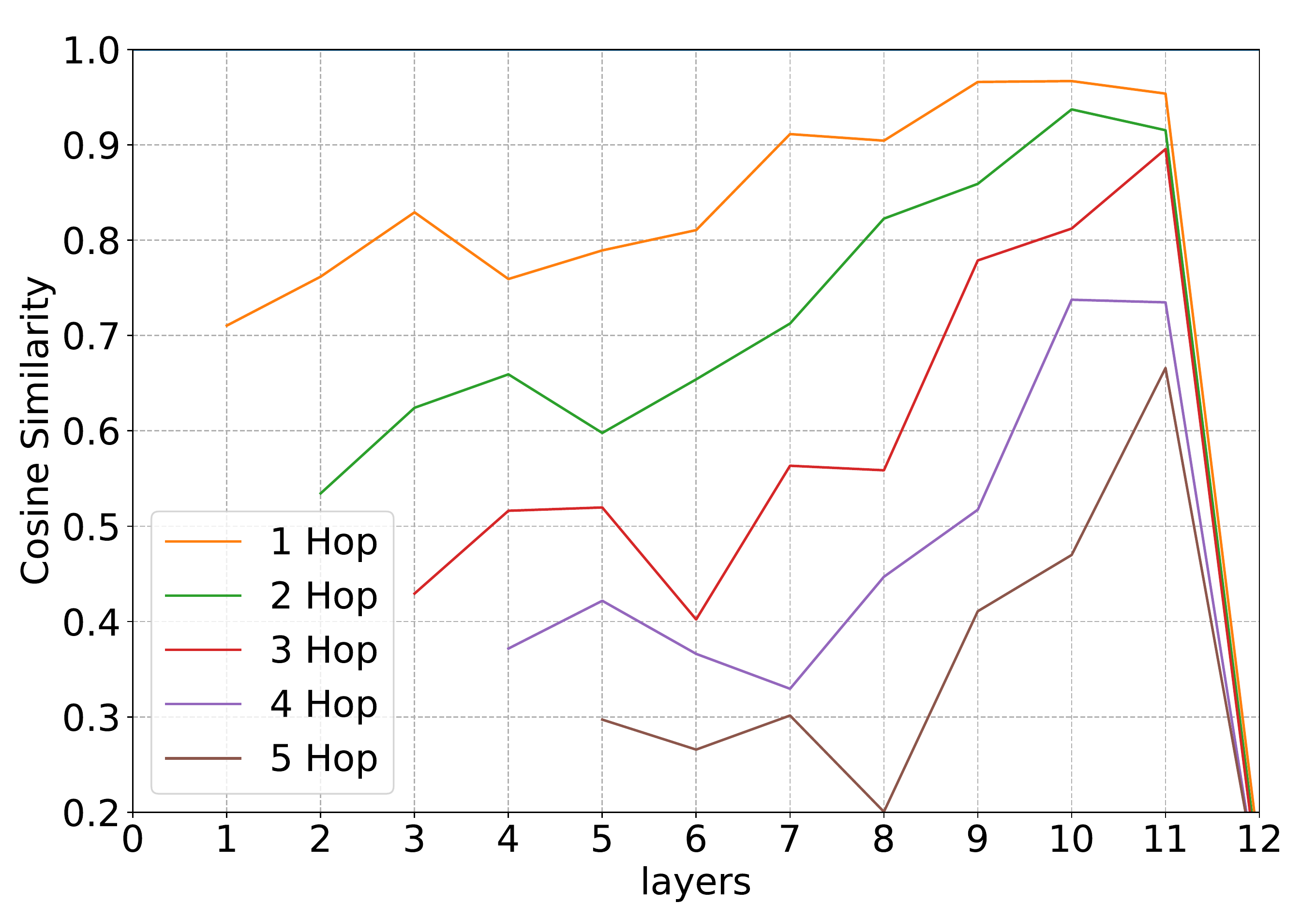}
        \caption{XLNET}
        \label{fig:q1_8}
    \end{subfigure}
\caption{Evolving word representation patterns across layers measured by
cosine similarity, where (a-d) show the similarity across layers and (e-h)
show the similarity over different hops. Four contextualized word 
representation models (BERT, SBERT, RoBERTa and XLNET) are tested.} \label{fig:q1}

\end{figure*}

\section{Word Representation Evolution across Layers} \label{sec:patten}
    
Although studies have been done in the understanding of the word
representation learned by deep contextualized models, none of them
examine how a word representation evolves across layers.  To observe 
such an evolving pattern, we design experiments in this section
by considering the following four BERT-based models.
\begin{itemize}
\item BERT \cite{BERT}. It employs the bi-directional training of the
transformer architecture and applies it to language modeling.
Unsupervised objectives, including the masked language model and the
next sentence prediction, are incorporated. 
\item SBERT \cite{reimers2019sentence}. It integrates the Siamese
network with a pre-trained BERT model. The supervised training objective on sentence pairs
is added to learn high quality sentence embedding. 
\item RoBERTa \cite{liu2019roberta}. It adapts the training process of
BERT to more general environments such as longer sequences, bigger
batches, more data and mask selection schemes, etc. The next sentence
prediction objective is removed. 
\item XLNET \cite{XLNET}. It adopts the Transformer-XL architecture, which
is trained with the Auto-Regressive (AR) objective. 
\end{itemize}
The above four BERT-based models have two variants; namely, the 12-layer
base model and the 24-layer large model. We choose their base models in
the experiments, which are pre-trained on their respective language
modeling tasks. 

To quantify the evolution of word representations across layers of deep
contextualized models, we measure the pair-wise cosine similarity
between 1- and $N$-hop neighbors. By the 1-hop neighbor, we refer to the
representation in the preceding or the succeeding layer of the current
layer. Generally, word $w$ has $(N+1)$ representations of dimension $d$
for a $N$-layer transformer network. The whole representation set for
$w$ can be expressed as
\begin{equation}
v_w^0, \, v_w^1, \cdots , v_w^N,
\end{equation}
where $v_w^i \in\mathbb{R}^{d}$ denotes the representation of word $w$ at the $i$-th layer.
The pair-wise cosine similarity between representations of the $i$-th and the $j$-th layers
can be computed as
\begin{equation} \label{eq:cossim}
\mathbf{CosSim}(i,j)=\frac{\langle v_w^i, v_w^j\rangle}{|v_w^i||v_w^j|}.
\end{equation}

To obtain statistical results, we extract word representations from all
sentences in the popular STS-Benchmark dataset \cite{cer2017semeval}.
The dataset contains $8628$ sentence pairs from three categories:
captions, news and forum. The similarity map is non-contextualized, which means we treat all words as isolated ones.  We
average the similarity map for all words to present the pattern for
contextualized word embedding models. 

\begin{table*}[htb]
\centering
\caption{Word groups based on the variance level. Less significant words 
in a sentence are underlined.}\label{tab:t1}
        \begin{adjustbox}{width=\textwidth,center}
         \begin{tabular}{||c | c | c | c||} 
         \hline
         Variance & Low  & Middle  & High  \\ [0.5ex] 
         \hline\hline
         \multirow{5}{*}{SBERT}  & \underline{can}, \underline{end}, \underline{do}, \underline{would}, time, \underline{all} ,say,                    & percent, security, \underline{mr}, \underline{into}, military,                        & eating, walking, small, room, person, \\ 
                                 & says, \underline{how}, \underline{before}, \underline{more}, east,                                            & \underline{she}, arms, \underline{they}, nuclear, head, billion,                      & children, grass, baby, cat, bike, field, \\
                                 & \underline{be}, \underline{have}, \underline{so}, \underline{could}, \underline{that}, \underline{than},       & \underline{on}, \underline{another}, \underline{around}, \underline{their}, million,  & runs, potato, horse, snow, ball, dogs, dancing \\
                                 & \underline{been}, south, united, \underline{what}, peace,                                                     & killed, mandela, arrested, wearing, three,                                            & men, dog, running, women, boy, jumping, \\
                                 & \underline{to}, states, against, \underline{since}, \underline{first}, \underline{last}                       & \underline{his}, \underline{her}, city, \underline{through}, cutting, green, oil      & plane, train, man, camera, woman, guitar \\ [1ex] 
         \hline
         \multirow{5}{*}{BERT}   & \underline{have}, \underline{his}, \underline{their}, \underline{last}, runs, \underline{would}               & jumping, \underline{on}, against, \underline{into}, man, baby                         & military, nuclear, killed, dancing, percent\\ 
                                 & \underline{been}, running, \underline{all}, \underline{than}, \underline{she}, \underline{that}               & \underline{around}, walking, person, green, \underline{her},                          & peace, plane, united, \underline{mr}, bike, guitar, \\
                                 & \underline{to}, cat, boy, \underline{be}, \underline{first}, woman, \underline{how}                           & \underline{end}, \underline{through}, \underline{another}, three, \underline{so},     & oil, train, children, arms, east, camera \\
                                 & cutting, \underline{since}, dogs, dog, say,                                                                   & wearing, mandela, south, \underline{do}, potato,                                      & grass, ball, field, room, horse, \underline{before}, billion \\
                                 & \underline{could}, \underline{more}, man, small, eating                                                       & \underline{they}, \underline{what}, women, says, \underline{can}, arrested            & city, security, million, snow, states, time \\ [1ex] 
        \hline
         \end{tabular}
         \end{adjustbox}
\end{table*}
            
Figs. \ref{fig:q1} (a)-(d) show the similarity matrix across layers for
four different models. Figs. \ref{fig:q1} (e)-(h) show the patterns
along the offset diagonal. In general, we see that the representations
from nearby layers share a large similarity value except for that in the
last layer. Furthermore, we observe that, except for the main diagonal,
offset diagonals do not have a uniform pattern as indicated by the blue
arrow in the associated figure. For BERT, SBERT and RoBERTa, the
patterns at intermediate layers are flatter as shown in Figs.
\ref{fig:q1} (e)-(g). The representations between consecutive layers
have a cosine similarity value that larger than 0.9. The rapid change
mainly comes from the beginning and the last several layers of the
network. This explains why the middle layers are more transferable to
other tasks as observed in \cite{liu2019linguistic}. Since the
representation in middle layers are more stable, more generalizable
linguistic properties are learned there.  As compared with BERT, SBERT
and RoBERTa, XLNET has a very different evolving pattern of word
representations.  Its cosine similarity curve as shown in Fig.
\ref{fig:q1} (h) is not concave.  This can be explained by the fact that
XLNET deviates from BERT significantly from architecture selection to
training objectives. It also sheds light on why SBERT
\cite{reimers2019sentence}, which has XLNET as the backbone for sentence
embedding generation, has sentence embedding results worse than BERT,
given that XLNET is more powerful in other NLP tasks. 

We see from Figs. \ref{fig:q1} (e)-(g) that the word representation
evolving patterns in the lower and the middle layers of BERT, SBERT and
RoBERTa are quite similar. Their differences mainly lie in the last
several layers. SBERT has the largest drop while RoBERTa has the minimum
change in cosine similarity measures in the last several layers.  SBERT
has the highest emphasis on the sentence-pair objective since it uses
the Siamese network for sentence pair prediction. BERT puts some focus
on the sentence-level objective via next-sentence prediction. In
contrast, RoBERTa removes the next sentence prediction completely in
training.  

We argue that faster changes in the last several layers are related to
the training with the sentence-level objective, where the distinct
sentence level information is reflected. Generally speaking, if more
information is introduced by a word, we should pay special attention to
its representation. To quantify such a property, we propose two metrics
(namely, alignment and novelty) in Sec. \ref{subsec:alignment}. 
    
We have so far studied the evolving pattern of word representations
across layers. We may ask whether such a pattern is word dependent. This
question can be answered below. As shown in Fig. \ref{fig:q1}, the
offset diagonal patterns are pretty similar with each other in the mean.
Without loss of generality, we conduct experiments on the offset-1
diagonal that contains 12 values as indicated by the arrow in Fig.
\ref{fig:q1}. We compute the variances of these 12 values to find the
variability of the 1-hop cosine similarity values with respect to
different words. The variance is computed for each word in BERT and
SBERT\footnote{Since RoBERTa and XLNET use a special tokenizer, which
cannot be linked to real word pieces, we do not test on RoBERTa and
XLNET here.}. We only report words that appear more than 50 times to
avoid randomness in Table \ref{tab:t1}.  The same set of words were
reported for BERT and SBERT models.  The words are split into three
categorizes based on their variance values. The insignificant words in a
sentence are underlined.  We can clearly see from the table that words
in the low variance group are in general less informative. In contrast,
words in the high variance group are mostly nouns and verbs, which
usually carry richer content. 

\begin{table}[htb]
\centering
\caption{Correlation coefficients and $p$-value between variance level and inverse document frequency (IDF).}\label{tab:q2_2}
        \begin{adjustbox}{width=0.5\columnwidth,center}
         \begin{tabular}{|c | c | c |} 
         \hline
         Model & $\rho$ & $p$-value  \\
         \hline
         BERT  & 31.89 & 3.85e-09\\ \hline
         SBERT & 20.62 & 1.87e-05\\ \hline
         \end{tabular}
         \end{adjustbox}
\end{table}

To further verify this phenomena, we compute the Spearman's rank correlation coefficients between the variance level and inverse document frequency measure. As showed in Table \ref{tab:q2_2}, a positive correlation between these two values are presented and the $p$-value also indicates a statistically significant correlation.

We conclude that more informative words in
deep contextualized models vary more while insignificant words vary
less.  This finding motivates us to design a module that can distinguish
important words in a sentence in Sec.  \ref{subsec:importance}. 

\section{Proposed SBERT-WK Method} \label{sec:dissecting_bert}
    
\begin{figure}[tb]
\centering
\includegraphics[width=0.5\textwidth]{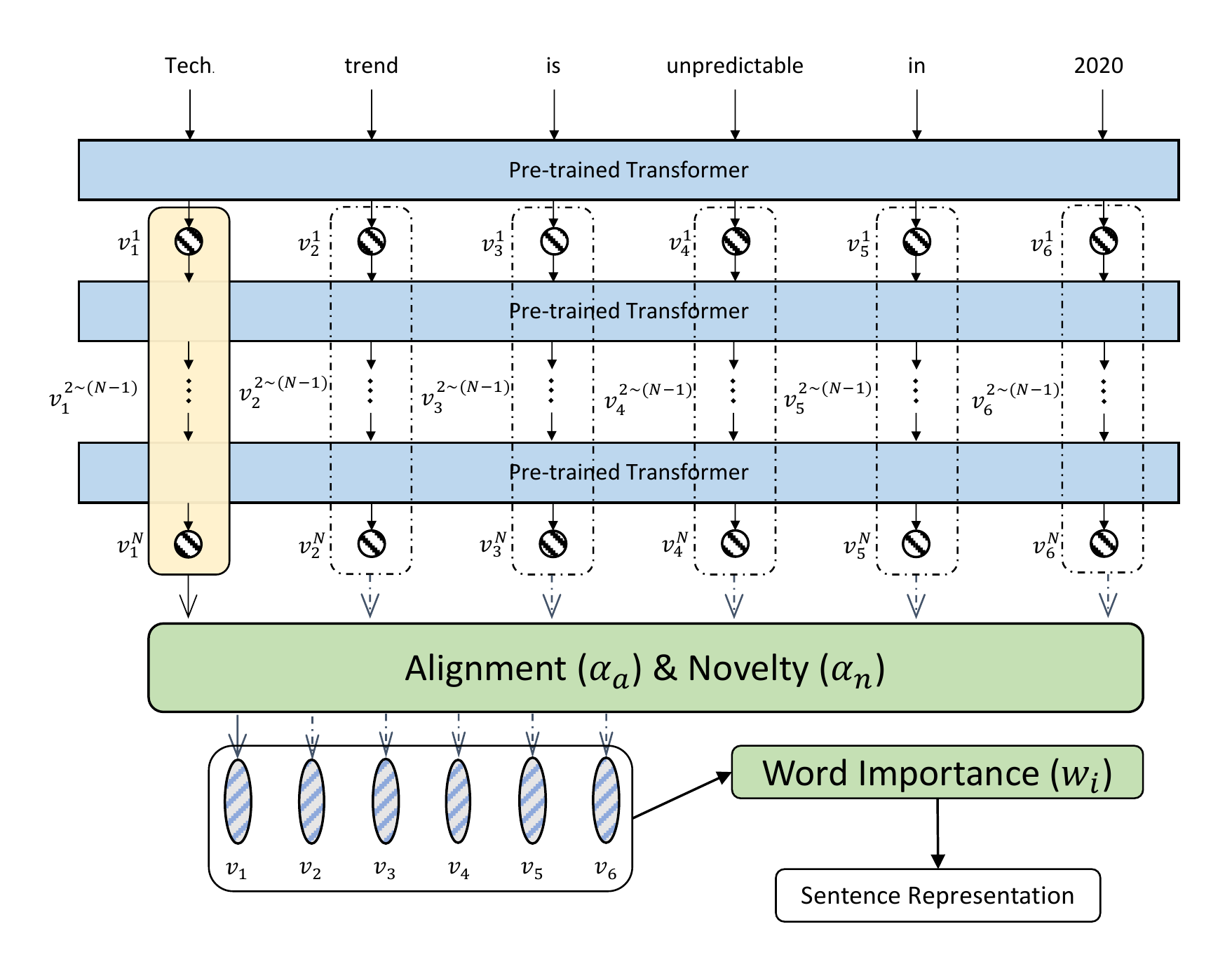}
\caption{Illustration for the proposed SBERT-WK model.}\label{fig:m1}
\end{figure}

We propose a new sentence embedding method called SBERT-WK in this section.
The block diagram of the SBERT-WK method is shown in Fig. \ref{fig:m1}.
It consists of the following two steps: 
\begin{enumerate}
\item Determine a unified word representation for each word in a
sentence by integrating its representations across layers by examining
its alignment and novelty properties. 
\item Conduct a weighted average of unified word representations based
on the word importance measure to yield the ultimate sentence embedding
vector. 
\end{enumerate}
They are elaborated in the following two subsections, respectively.

\subsection{Unified Word Representation Determination }\label{subsec:alignment}

As discussed in Sec. \ref{sec:patten}, the word representation evolves
across layers.  We use $v_w^i$ to denote the representation of word $w$
at the $i$th layer.  To determine the unified word representation,
$\hat{v}_w$, of word $w$ in Step 1, we assign weight $\alpha_i$ to its
$i$th layer representation, $v_w^i$, and take an average:
\begin{equation}\label{eq:uwr}
\hat{v}_w = \sum_{i=0}^N \alpha (v_w^i) v_w^i,
\end{equation}
where weight $\alpha$ can be derived based on the inverse alignment and the
novelty two properties.

\subsubsection{Inverse Alignment Measure}

We define the (layer-wise) neighboring matrix of $v_w^i$ as
\begin{equation}\label{eq:context}
\mathbf{C}=[v_w^{i-m}, \cdots , v_w^{i-1}, v_w^{i+1}, \cdots  v_w^{i+m}]
\in \mathbb{R}^{d \times 2m},
\end{equation}
where $d$ is the word embedding dimension and $m$ is the neighboring window
size.  We can compute the pair-wise cosine similarity between $v_w^i$ and
all elements in the neighboring window $C(v_w^i)$ and use their average to
measure how $v_w^i$ aligns with the neighboring word vectors. 
Then, the alignment similarity score of $v_w^i$ can be defined as
\begin{equation}\label{eq:align}
\beta_{a}(v_w^i)=\frac{1}{2m}\sum_{j=i-m,j\neq i}^{i+m}
\frac{\langle v_w^i, v_w^j\rangle}{|v_w^i||v_w^j|}.
\end{equation}
If a word representation at a layer aligns well with its neighboring word
vectors, it does not provide much additional information. Since it is
less informative, we can give it a smaller weight. Thus, we use the
inverse of the alignment similarity score as the weight for word $w$ 
at the $i$-th layer. Mathematically, we have
\begin{equation}\label{eq:align2}
\alpha_{a}(v_w^i)=\frac{K_a}{\beta_{a}(v_w^i)},
\end{equation}
where $K_a$ is a normalization constant independent of $i$ and it is 
chosen to normalize the sum of weights:
$$
\sum_{i=1}^N \alpha_{a}(v_w^i) = 1.
$$
We call $\alpha_{a}(v_w^i)$ the inverse alignment weight.

\subsubsection{Novelty Measure}

Another way to measure the new information of word representation
$v_w^i$ is to study the new information brought by it with respect to
the subspace spanned words in its neighboring window.  Clearly, words in the matrix $\mathbf{C}$ form a subspace. We can decompose $v_w^i$
into two components: one contained by the subspace and the other
orthogonal to the subspace. We view the orthogonal one as its novel
component and use its magnitude as the novelty score. 
By singular value decomposition (SVD), we can factorize matrix $\mathbf{M}$ of
dimension $m\times n$ into the form $\mathbf{M}=\mathbf{U\Sigma V}$,
where $\mathbf{U}$ is an $m\times n$ matrix with orthogonal columns,
$\mathbf{\Sigma}$ is an $n\times n$ diagonal matrix with non-negative
numbers on the diagonal and $\mathbf{V}$ is $n\times n$ orthogonal
matrix. First, we decompose the matrix
$\mathbf{C}$ in Eq. (\ref{eq:context}) to $\mathbf{C}=\mathbf{U\Sigma
V}$ to find the orthogonal basis for the neighboring words. The orthogonal
column basis for $\mathbf{C}$ is represented by matrix $\mathbf{U}$.
Thus, the orthogonal component of $v_w^i$ with respect to $\mathbf{C}$
can be computed as
\begin{equation}\label{eq:new_dir}
q_w^i=v_w^i - \mathbf{U}\mathbf{U^T} v_w^i.
\end{equation}
The novelty score of $v_w^i$ is computed by 
\begin{equation}\label{eq:novel}
\alpha_n(v_w^i)=\frac{K_n ||q_w^i||_2}{||v_w^i||_2},
\end{equation}
where $K_n$ is a normalization constant independent of $i$ 
and it is chosen to normalize the sum of weights:
$$
\sum_{i=1}^N \alpha_{n}(v_w^i) = 1.
$$
We call $\alpha_{n}(v_w^i)$ the novelty weight.

\subsubsection{Unified Word Representation}

We examine two ways to measure the new information brought by word
representation $v_w^i$ at the $i$-th layer. We may consider a
weighted average of the two in form of
\begin{gather}\label{eq:alpha}
\alpha_c (v_w^i, \omega) = \omega \alpha_a (v_w^i)+ (1-\omega) \alpha_n (v_w^i),
\end{gather}
where $0 \le \omega \le 1$ and $\alpha_c (v_w^i, \omega)$ is called the
combined weight. We compare the performance of three cases (namely,
novelty weight $\omega=0$, inverse alignment weight $\omega=1$ and
combined weight $\omega=0.5$) in the experiments.  A unified word
representation is computed as a weighted sum of its representations in
different layers:
\begin{gather}\label{eq:new}
\hat{v}_w = \sum_{i=0}^N \alpha_c (v_w^i) v_w^i.
\end{gather}
We can view $v_w$ as the new contextualized word 
representation for word $w$. 

\subsection{Word Importance}\label{subsec:importance}

As discussed in Sec. \ref{sec:patten}, the variances of the pair-wise
cosine-similarity matrix can be used to categorize words into different
groups.  Words of richer information usually have a larger variance.  By
following the line of thought, we can use the same variance to determine
the importance of a word and merge multiple words in a sentence to
determine the sentence embedding vector. This is summarized below.

For the $j$-th word in a sentence denoted by $w(j)$, we first compute its
cosine similarity matrix using its word representations from all layers
as shown in Eq.  (\ref{eq:cossim}). Next, we extract the offset-1
diagonal of the cosine similarity matrix, compute the variance of the
offset-1 diagonal values and use $\sigma_j^2$ to denote the variance of
the $j$th word. Then, the final sentence embedding ($v_s$) can be
expressed as
\begin{equation}\label{eq:sen_emb}
v_s=\sum_j \omega_j \hat{v}_{w(j)}, 
\end{equation}
where $\hat{v}_{w(j)}$ is the the new contextualized word representation
for word $w(j)$ as defined in Eq. (\ref{eq:new}) and
\begin{equation}\label{eq:sen_omega}
\omega_j = \frac{|\sigma_j^2|}{\sum_k |\sigma_k^2|}.
\end{equation}
Note that the weight for each word is the $l_1$-normalized variance as
shown in Eq. (\ref{eq:sen_omega}).  To sum up, in our sentence embedding
scheme, words that evolve faster across layers with get higher weights
since they have larger variances. 

\subsection{Computational Complexity} \label{subsec:computation_complexity}

The main computational burden of SBERT-WK comes from the SVD
decomposition, which allows more fine-grained analysis in novelty
measure.  The context window matrix $\mathbf{C}$ is decomposed into the
product of three matrices $\mathbf{C}=\mathbf{U\Sigma V}$. The
orthogonal basis is given by matrix $\mathbf{U}$. The context window
matrix is of size $d\times 2m$, where $d$ is the word embedding size and
$2m$ is the whole window size. In our case, $d$ is much larger than $m$
so that the computational complexity for SVD is $O(8dm^2)$, where
several terms are ignored. 

Instead of performing SVD decomposition, we use the QR factorization in
our experiments as an alternative because of its computational
efficiency. With QR factorization, we first concatenate the center word
vector represenation $v_w^i$ to the context window matrix $\mathbf{C}$
to form a new matrix
\begin{equation}\label{eq:NOV1}
\tilde{\mathbf{C}}=[v_w^{i-m}, \cdots , v_w^{i-1}, v_w^{i+1}, \cdots ,
v_w^{i+m}, v_w^i]\in \mathbb{R}^{d\times(2m+1)}
\end{equation}
has $2m+1$ word representations. We perform the QR factorization on
$\tilde{\mathbf{C}}$, and obtain $\tilde{\mathbf{C}} =
\mathbf{Q}\mathbf{R}$, where non-zero columns of matrix
$\mathbf{Q}\in\mathbb{R}^{d\times (2m+1)}$ are orthonormal basis and
$\mathbf{R}\in\mathbb{R}^{(2m+1)\times (2m+1)}$ is an upper triangular
matrix that contains the weights for word representations under the
basis of $\mathbf{Q}$. 
We denote the $i$th column of $\mathbf{Q}$ and $\mathbf{R}$ as
$\mathbf{q}^i$ and $\mathbf{r}^i$, respectively. With QR factorization,
$\mathbf{r}^{2m+1}$ is the representation of $v_w^i$ under
the orthogonal basis formed by matrix $\mathbf{Q}$. The new direction
introduced to the context by $v_w^i$ is represented as $q^{2m+1}$.
Then, the last component of $r^{2m+1}$ is the weight for the new
direction, which is denoted by $r^{2m+1}_{-1}$. Then, the novelty 
weight can be derived as:
\begin{equation}\label{eq:NOV}
\alpha_n (v_w^i) =\frac{K_n r^{2m+1}_{-1}}{|r^{2m+1}|},
\end{equation}
where $K_n$ is the normalization constant.  The inverse alignment weight
can also computed under the new basis $\mathbf{Q}$. 

The complexity of the QR factorization is $O(d(2m+1)^2)$, which is two
times faster than the SVD decomposition. In practice, we see little
performance difference between these two methods. The experimental
runtime is compared in Sec. \ref{exp:computation}

\section{Experiments} \label{sec:experiments}

Since our goal is to obtain a general purpose sentence embedding method,
we evaluate SBERT-WK on three kinds of evaluation tasks. 
\begin{itemize}
\item Semantic textual similarity tasks. \\
They predict the similarity between two given sentences. They can be
used to indicate the embedding ability of a method in terms of
clustering and information retrieval via semantic search. 
\item Supervised downstream tasks. \\
They measure embedding's transfer capability to downstream tasks
including entailment and sentiment classification. 
\item Probing tasks. \\
They are proposed in recent years to measure the linguistic features 
of an embedding model and provide fine-grained analysis. 
\end{itemize}

These three kinds of evaluation tasks can provide a comprehensive test
on our proposed model.  The popular SentEval toolkit
\cite{conneau2018senteval} is used in all experiments. The proposed
SBERT-WK method can be built upon several state-of-the-art pre-trained
language models including BERT, RoBERTa and XLNET. Here, we
evaluate it on top of two models:
BERT 
and 
RoBERTa. Both pre-trained models are been further fine-tuned with natural language inference data as described in \cite{reimers2019sentence}. We adopt their base models that contain 12 transformer layers as well as large models with 24 layers. 

For performance benchmarking, we compare SBERT-WK with the following 
10 different methods, including parameterized and non-parameterized models. 
\begin{enumerate}
\item Average of GloVe word embeddings; 
\item Average the last layer token representations of BERT;
\item Use [CLS] embedding from BERT, where [CLS] is used for
next sentence prediction in BERT; 
\item SIF model \cite{arora2016simple}, which is a non-parameterized model 
that provides a strong baseline in textual similarity tasks;
\item GEM model \cite{yang2019parameter}, which is a non-parameterized model deriving from the analysis of static word embedding space;
\item $p$-mean model \cite{ruckle2018concatenated} that incorporates
multiple word embedding models; 
\item Skip-Thought \cite{kiros2015skip}; 
\item InferSent \cite{conneau2017supervised} with both GloVe and FastText
versions; 
\item Universal Sentence Encoder \cite{cer2018universal}, which is a
strong parameterized sentence embedding using multiple objectives and
transformer architecture; 
\item SBERT, which is a state-of-the-art sentence embedding 
model by training the Siamese network over BERT. 
\end{enumerate}

\subsection{Semantic Textural Similarity}

To evaluate semantic textual similarity, we use 2012-2016 STS datasets
\cite{agirre2012semeval, agirre2013semeval, agirre2014semeval,
agirre2015semeval, agirre2016semeval}.  They contain sentence pairs and
labels between 0 and 5, which indicate their semantic relatedness.  Some
methods learn a complex regression model that maps sentence pairs to
their similarity score. Here, we use the cosine similarity between
sentence pairs as the similarity score and report both Pearson and Spearman's rank
correlation coefficient. More details of these datasets can be found in \cite{conneau2018senteval}. 

Semantic relatedness is a special kind of similarity task, and we use
the SICK-R \cite{marelli2014sick} and the STS Benchmark dataset
\cite{cer2017semeval} in our experiments. Being different from
STS12-STS16, the semantic relatedness datasets are under the supervised
setting where we learn to predict the probability distribution of
relatedness scores.  The STS Benchmark dataset is a popular dataset to
evaluate supervised STS systems. It contains 8,628 sentences from three
categories (captions, news and forums) and they are divided into train
(5,749), dev (1,500) and test (1,379).

In our experiments, we do not include the representation from the first
three layers since their representations are less contextualized as
reported in \cite{ethayarajh2019contextual}. Some superficial
information is captured by those representations and they play a
subsidiary role in most tasks \cite{BERT_STRU}. We set the context
window size to $m=2$ in all evaluation tasks.

            \begin{table*}[htb]
                \centering
                \caption{Experimental results on various textual similarity tasks in terms of the Pearson 
                correlation coefficients (left, $\times 100$) and Spearman's rank correlation coefficients (right, $\times 100$), where the best results are shown in bold face.}\label{exp:STS}                \resizebox{2\columnwidth}{!}{
                \begin{tabular}{ c | c | c | c | c | c | c | c | c || c} 
                    \hline
                    \textbf{Model} & \textbf{Dim} & \textbf{STS12} & \textbf{STS13} & \textbf{STS14} & \textbf{STS15} & \textbf{STS16} & \textbf{STSB} & \textbf{SICK-R} & \textbf{Avg.} \\ \hline\hline
                    \multicolumn{3}{l}{\textit{Non-Parameterized models}}\\
                    \hline\hline
                    Avg. GloVe embeddings & 300 & 52.3 / 53.3 & 50.5 / 50.7 & 55.2 / 55.6 & 56.7 / 59.2 & 54.9 / 57.7 & 65.8 / 62.8 & 80.0 / 71.8 & 59.3 / 58.7 \\
                    SIF (Arora et al., 2017) & 300 & 56.2 / \ \ -\ \ \  & 56.6 / \ \ -\ \ \ & 68.5 / \ \ -\ \ \ & 71.7 / \ \ -\ \ \ & \ \ -\ \ \ / \ \ -\ \ \ & 72.0 / \ \ -\ \ \ & 86.0 / \ \ -\ \ \ & 68.5 / \ \ -\ \ \ \\
                    $p$-mean (Rucklle et al., 2018) & 3600 & 54.0 / \ \ -\ \ \ & 52.0 / \ \ -\ \ \ & 63.0 / \ \ -\ \ \ & 66.0 / \ \ -\ \ \ & 67.0 / \ \ -\ \ \ & 72.0 / \ \ -\ \ \ & 86.0 / \ \ -\ \ \ & 65.7 / \ \ -\ \ \ \\
                    \hline\hline
                    \multicolumn{3}{l}{\textit{Parameterized models}}\\
                    \hline\hline                    
                    Skip-Thought (Kiros et al., 2015) & 4800 & 41.0 / \ \ -\ \ \ & 29.8 / \ \ -\ \ \ & 40.0 / \ \ -\ \ \ & 46.0 / \ \ -\ \ \ & 52.0 / \ \ -\ \ \ & 75.0 / \ \ -\ \ \ & 86.0 / \ \ -\ \ \ & 52.8 / \ \ -\ \ \ \\
                    InferSent-GloVe (Conneau et al., 2017) & 4096 & 59.3 / 60.3 & 58.8 / 58.7 & 69.6 / 66.7 & 71.3 / 72.2 & 71.5 / 72.6 & 75.7 / 75.3 & \textbf{88.4} / 82.5 & 70.7 / 69.8 \\
                    USE  (Cer et al., 2018) & 512 & 61.4 / 62.0 & 63.5 / 64.2 & 70.6 / 67.0 & 74.3 / 75.9 & 73.9 / \textbf{77.3} & 78.2 / 77.1 & 85.9 / 79.8 & 72.5 / 71.9 \\
                    BERT [CLS] (Devlin et al., 2018) & 768 & 27.5 / 32.5 & 22.5 / 24.0 & 25.6 / 28.5 & 32.1 / 35.5 & 42.7 / 51.1 & 52.1 / 51.8 & 70.0 / 64.8 & 38.9 / 41.2 \\
                    Avg. BERT embedding (Devlin et al., 2018) & 768 & 46.9 / 50.1 & 52.8 / 52.9 & 57.2 / 54.9 & 63.5 / 63.4 & 64.5 / 64.9 & 65.2 / 64.2 & 80.5 / 73.5 & 61.5 / 60.6 \\
                    SBERT-base (Reimers et al., 2019) & 768 & 64.6 / 63.8 & 67.5 / 69.3 & 73.2 / 72.9 & 74.3 / 75.2 & 70.1 / 73.3 & 74.1 / 74.5 & 84.2 / 79.3 & 72.5 / 72.6\\
                    SBERT-large (Reimers et al., 2019) & 1024 & 66.9 / 66.8 & \textbf{69.4} / \textbf{71.4} & 74.2 / \textbf{74.3} & \textbf{77.2} / \textbf{78.2} & 72.8 / 75.7 & 75.6 / 75.8 & 84.7 / 80.3 & 74.4 / 74.6 \\
                    \hline\hline
                    SBERT-WK-base & 768 & \textbf{70.2} / \textbf{68.2} & 68.1 / 68.8 & \textbf{75.5} / \textbf{74.3} & 76.9 / 77.5 & \textbf{74.5} / 77.0 & 80.0 / 80.3 & 87.4 / 82.3 & \textbf{76.1} / \textbf{75.5} \\
                    SRoBERTa-WK-base & 768 & 68.4 / 67.6 & 63.9 / 65.9 & 71.5 / 72.8 & 67.9 / 75.2 & 70.2 / 74.0 & \textbf{80.7} / \textbf{81.1} & 87.6 / \textbf{82.9} & 72.9 / 74.2 \\
                    \hline
                \end{tabular}}
            \end{table*}


The results are given in Table \ref{exp:STS}. We see that the use of
BERT outputs directly generates rather poor performance.  For example,
the [CLS] token representation gives an average correlation score of
38.9/41.2 only. Averaging BERT embeddings provides an average correlation
score of 61.5/60.6. This is used as the default setting of generating
sentence embedding from BERT in the bert-as-service toolkit
\footnote{https://github.com/hanxiao/bert-as-service}. They are both
worse than non-parameterized models such as SIF, which is using static word embedding. Their poor
performance could be partially attributed to that the model is not trained using a
similar objective function. The masked language model and next sentence
prediction objectives are not suitable for a linear integration of
representations. The study in \cite{ethayarajh2018towards} explains how
linearity is exploited in static word embeddings (e.g., word2vec) and it
sheds light on contextualized word representations as well.  Among the
above two methods, we recommend averaging BERT outputs because it
captures more inherent structure of the sentence while the [CLS] token
representation is more suitable for some downstream classification tasks
as shown in Table \ref{exp:downstream_result}. 

We see from Table \ref{exp:STS} that InferSent, USE and SBERT provide
the state-of-the-art performance on textual similarity tasks.
Especially, InferSent and SBERT have a mechanism to incorporate the
joint representation of two sentences such as the point-wise difference
or the cosine similarity. Then, the training process learns the
relationship between sentence representations in a linear manner and
compute the correlation using the cosine similarity, which is a perfect
fit. Since the original BERT model is not trained in this manner, the
use of the BERT representation directly would give rather poor
performance.  The similar phenomena happens to other BERT-based models as well. Therefore, BERT-based models are desired to be fine-tuned with sentence pairs before evaluating with cosine similarities.

As compared with other methods, SBERT-WK improves the performance on
textual similarity tasks by a significant margin. It is worthwhile to
emphasize that we use only 768-dimension vectors for sentence embedding
while InferSent uses 4096-dimension vectors. As explained in
\cite{ruckle2018concatenated,eger2019pitfalls,conneau2017supervised},
the increase in the embedding dimension leads to increased performance
for almost all models. This may explain SBERT-WK is slightly inferior to InferSent on the SICK-R dataset. For all other tasks, SBERT-WK achieves substantial better performance even with a smaller embedding size.

While RoBERTa can supersede BERT model in supervised tasks, we did not witness obvious improvement on STS datasets. During the model pre-training stage, unlike BERT, RoBERTa is not incorporating any sentence-level objective. That may empower RoBERTa with less sentence level information across layers.

On STS dataset, we also tested the large model (24 layers) but general led to worse result than the base model. We would consider the large model may need different hyperparameter settings or specific fine-tune schemes in order to perform well on STS tasks. However, even our model only contained 12 layers, it can still outperform the 24 layer model used in SBERT.

\subsection{Supervised Downstream Tasks}

For supervised tasks, we compare SBERT-WK with other sentence embedding
methods in the following eight downstream tasks. 
\begin{itemize}
\item MR: Binary sentiment prediction on movie reviews \cite{pang2005seeing}.
\item CR: Binary sentiment prediction on customer product reviews \cite{hu2004mining}.
\item SUBJ: Binary subjectivity prediction on movie reviews and plot summaries \cite{pang2004sentimental}.
\item MPQA: Phrase-level opinion polarity classification \cite{wiebe2005annotating}.
\item SST2: Stanford Sentiment Treebank with binary labels \cite{socher2013recursive}.
\item TREC: Question type classification with 6 classes \cite{li2002learning}.
\item MRPC: Microsoft Research Paraphrase Corpus for paraphrase prediction \cite{dolan2004unsupervised}.
\item SICK-E: Natural language inference dataset \cite{marelli2014sick}.
\end{itemize}
More details on these datasets can be found in \cite{conneau2018senteval}. 

The design of our sentence embedding model targets at the transfer
capability to downstream tasks. Typically, one can tailor a pre-trained
language model to downstream tasks through tasks-specific fine-tuning.
It was shown in previous work \cite{arora2016simple,yang2019parameter}, that subspace analysis methods are more
powerful in semantic similarity tasks. However, we would like to show
that sentence embedding can provide an efficient way for downstream
tasks as well. In particular, we demonstrate that SBERT-WK does not hurt
the performance of pre-trained language models. Actually, it can even
perform better than the original model in downstream tasks under both BERT and RoBERTa backbone settings. 

For SBERT-WK, we use the same setting as the one in semantic similarity
tasks. For downstream tasks, we adopt a multi-layer-perception (MLP)
model that contains one hidden layer of 50 neurons. The batch size is
set to 64 and the Adam optimizer is adopted in the training. All
experiments are trained with 4 epochs.  For MR, CR, SUBJ, MPQA and MRPC,
we use the nested 10-fold cross validation.  For SST2, we use the
standard validation. For TREC and SICK-E, we use the cross validation. 

            \begin{table*}[htb]
                \centering
                \caption{Experimental results on eight supervised downstream tasks, where the best results 
                 are shown in bold face.}\label{exp:downstream_result}
                \resizebox{2\columnwidth}{!}{
                \begin{tabular}{ c | c | c | c | c | c | c | c | c | c || c } 
                    \hline
                    \textbf{Model} & \textbf{Dim} & \textbf{MR} & \textbf{CR} & \textbf{SUBJ} & \textbf{MPQA} & \textbf{SST2} & \textbf{TREC} & \textbf{MRPC} & \textbf{SICK-E} & \textbf{Avg.} \\ \hline\hline
                    \multicolumn{3}{l}{\textit{Non-Parameterized models}}\\
                    \hline\hline
                    Avg. GloVe embeddings & 300 & 77.9 & 79.0 & 91.4 & 87.8 & 81.4 & 83.4 & 73.2 & 79.2 & 81.7 \\
                    SIF (Arora et al., 2017) & 300 & 77.3 & 78.6 & 90.5 & 87.0 & 82.2 & 78.0 & - & 84.6 & 82.6 \\
                    $p$-mean (Rucklle et al., 2018) & 3600 & 78.3 & 80.8 & 92.6 & 73.2 & 84.1 & 88.4 & 73.2 & 83.5 & 81.8 \\
                    GEM (Yang et al., 2019) & 900 & 79.8 & 82.5 & 93.8 & 89.9 & 84.7 & 91.4 & 75.4 & 86.2 & 85.5\\
                    \hline\hline
                    \multicolumn{3}{l}{\textit{Parameterized models}}\\
                    \hline\hline                    
                    Skip-Thought (Kiros et al., 2015) & 4800 & 76.6 & 81.0 & 93.3 & 87.1 & 81.8 & 91.0 & 73.2 & 84.3 & 83.5 \\
                    InferSent-GloVe (Conneau et al., 2017) & 4096 & 81.8 & 86.6 & 92.5 & 90.0 & 84.2 & 89.4 & 75.0 & \textbf{86.7} & 85.8 \\
                    Universal Sentence Encoder (Cer et al., 2018) & 512 & 80.2 & 86.0 & 93.7 & 87.0 & 86.1 & \textbf{93.8} & 72.3 & 83.3 & 85.3 \\
                    BERT [CLS] vector (Devlin et al., 2018) & 768 & 82.3 & 86.9 & \textbf{95.4} & 88.3 & 86.9 & \textbf{93.8} & 72.1 & 73.8 & 84.9\\
                    Avg. BERT embedding (Devlin et al., 2018) & 768 & 81.7 & 86.8 & 95.3 & 87.8 & 86.7 & 91.6 & 72.5 & 78.2 & 85.1 \\
                    SBERT-base (Reimers et al., 2019) & 768 & 82.4 & 88.9 & 93.9 & 90.1 & 88.4 & 86.4 & 75.5 & 82.3 & 86.0\\
                    SBERT-large (Reimers et al., 2019) & 1024 & 84.8 & 90.5 & 94.7 & 90.6 & 91.0 & 88.2 & 76.9 & 82.1 & 87.3\\
                    \hline\hline
                    SBERT-WK-base & 768 & 83.0 & 89.1 & 95.2 & \textbf{90.6} & 89.2 & 93.2 & 77.4 & 85.5 & 87.9 \\
                    SBERT-WK-large & 1024 & 85.2 & \textbf{91.6} & 95.2 & 90.7 & 91.8 &92.4 & 77.3 & 85.1 & 88.7\\
                    SRoBERTa-WK-base & 768 & 85.8 & 91.4 & 94.5 & 89.7 & 92.3 & 91.0 & \textbf{78.8} & 86.5 &  \textbf{88.8} \\
                    SRoBERTa-WK-large & 1024 & \textbf{87.2} & 91.3 & 94.2 & 91.1 & \textbf{93.2} & 88.0 & 77.9 & 82.1 & 88.2  \\
                    \hline
                \end{tabular}}
            \end{table*}

The experimental results on the eight supervised downstream tasks are given
in Table \ref{exp:downstream_result}.  Although it is desired to
fine-tune deep models for downstream tasks, we see that SBERT-WK still
achieves good performance without any fine-turning. As compared with the
other 10 benchmarking methods, SBERT-WK has the best performance in 5
out of the 8 tasks. For the remaining 3 tasks, it still ranks among the
top three. Unlike STS tasks, SRoBERTa-WK-base achieves the best
averaged performance (88.8\%) on supervised tasks.  For TREC,
SBERT-WK is inferior to the two best models, USE and BERT [CLS], by
0.6\%.  For comparison, the baseline SBERT is much worse than USE, and
SBERT-WK-base outperforms SBERT-base by 6.8\%. USE is particularly suitable TREC
since it is pre-trained on question answering data, which is highly
related to the question type classification task. In contrast, SBERT-WK
is not trained or fine-tuned on similar tasks. For SICK-E, SBERT-WK is
inferior to two InferSent-based methods by 1.2\%, which could be
attributed to the much larger dimension of InferSent. 

We observe that averaging BERT outputs and [CLS] vectors give pretty
similar performance. Although [CLS] provides poor performance for semantic
similarity tasks, [CLS] is good at classification tasks. This is because
that the classification representation is used in its model training.
Furthermore, the use of MLP as the inference tool would allow certain
dimensions to have higher importance in the decision process.  The
cosine similarity adopted in semantic similarity tasks treats all
dimension equally. As a result, averaging BERT outputs and [CLS] token
representation are not suitable for semantic similarity tasks.  If we
plan to apply the [CLS] representation and/or averaging BERT embeddings to
semantic textual similarity, clustering and retrieval tasks, we need to
learn an additional transformation function with external resources. 

\subsection{Probing Tasks}

It is difficult to infer what kind of information is present in sentence
representation based on downstream tasks.  Probing tasks focus more on
language properties and, therefore, help us understand sentence
embedding models. We compare SBERT-WK-base on 10 probing tasks so as to cover
a wide range of aspects from superficial properties to deep semantic
meanings. They are divide into three types \cite{conneau2018you}: 1)
surface information, 2) syntactic information and 3) semantic
information. 

            \begin{itemize}
                \item Surface Information
                \begin{itemize}
                    \item SentLen: Predict the length range of the input sentence with 6 classes.
                    \item WC: Predict which word is in the sentence given 1000 candidates.
                \end{itemize}
                \item Syntactic Information
                \begin{itemize}
                    \item TreeDepth: Predict depth of the parsing tree.
                    \item TopConst: Predict top-constituents of parsing tree within 20 classes.
                    \item BShift: Predict whether a bigram has been shifted or not.
                \end{itemize}
                \item Semantic Information
                \begin{itemize}
                    \item Tense: Classify the main clause tense with past or present.
                    \item SubjNum: Classify the subject number with singular or plural.
                    \item ObjNum: Classify the object number with singular or plural.
                    \item SOMO: Predict whether the noun/verb has been replaced by another one with the same part-of-speech character.
                    \item CoordInv: Sentences are made of two coordinate clauses. Predict whether it is inverted or not.
                \end{itemize}
            \end{itemize}

We use the same experimental setting as that used for supervised tasks.
The MLP model has one hidden layer of 50 neurons. The batch size is set
to 64 while Adam is used as the optimizer.  All tasks are trained in 4
epochs. The standard validation is employed.  Being Different from the
work in \cite{perone2018evaluation} that uses logistic regression for
the WC task in the category of surface information, we use the same MLP
model to provide simple yet fair comparison. 

            \begin{table*}[htb]
                \centering
                \caption{Experimental results on 10 probing tasks, where the best results are shown in bold face.}\label{exp:probing_result}                \resizebox{2\columnwidth}{!}{
                \begin{tabular}{ c | c | c | c | c | c | c | c | c | c | c | c } 
                    \hline
                    \multicolumn{2}{l|}{} & \multicolumn{2}{c|}{Surface} & \multicolumn{3}{c|}{Syntactic} & \multicolumn{5}{c}{Semantic} 
                    \\ \hline
                    \textbf{Model} & \textbf{Dim} & \textbf{SentLen} & \textbf{WC} & \textbf{TreeDepth} & \textbf{TopConst} & \textbf{BShift} & \textbf{Tense} & \textbf{SubjNum} & \textbf{ObjNum} & \textbf{SOMO} & \textbf{CoordInv}\\ \hline\hline
                    \multicolumn{3}{l}{\textit{Non-Parameterized models}}\\
                    \hline\hline
                    Avg. GloVe embeddings & 300 & 71.77 & 80.61 & 36.55 & 66.09 & 49.90 & 85.33 & 79.26 & 77.66 & 53.15 & 54.15 \\
                    $p$-mean (Rucklle et al., 2018) & 3600 & 86.42 & \textbf{98.85} & 38.20 & 61.66 & 50.09 & 88.18 & 81.73 & 83.27 & 53.27 & 50.45 \\
                    \hline\hline
                    \multicolumn{3}{l}{\textit{Parameterized models}}\\
                    \hline\hline                    
                    Skip-Thought (Kiros et al., 2015) & 4800 & 86.03 & 79.64 & 41.22 & \textbf{82.77} & 70.19 & \textbf{90.05} & 86.06 & 83.55 & 54.74 & 71.89 \\
                    InferSent-GloVe (Conneau et al., 2017) & 4096 & 84.25 & 89.74 & 45.13 & 78.14 & 62.74 & 88.02 & 86.13 & 82.31 & 60.23 & 70.34 \\
                    Universal Sentence Encoder (Cer et al., 2018) & 512 & 79.84 & 54.19 & 30.49 & 68.73 & 60.52 & 86.15 & 77.78 & 74.60 & 58.48 & 58.19 \\
                    BERT [CLS] vector (Devlin et al., 2018) & 768 & 68.05 & 50.15 & 34.65 & 75.93 & 86.41 & 88.81 & 83.36 & 78.56 & 64.87 & \textbf{74.32} \\
                    Avg. BERT embedding (Devlin et al., 2018) & 768 & 84.08 & 61.11 & 40.08 & 73.73 & \textbf{88.80} & 88.74 & 85.82 & 82.53 & \textbf{66.76} & 72.59 \\
                    SBERT-base (Reimers et al., 2019) & 768 & 75.55 & 58.91 & 35.56 & 61.49 & 77.93 & 87.32 & 79.76 & 78.40 & 62.85 & 65.34 \\
                    \hline\hline
                    SBERT-WK-base & 768 & \textbf{92.40} & 77.50 & \textbf{45.40} & 79.20 & 87.87 & 88.88 & 86.45 & \textbf{84.53} & 66.01 & 71.87 \\
                    \hline
                \end{tabular}}
            \end{table*}

The performance is shown in Table \ref{exp:probing_result}. We see that
SBERT-WK yields better results than SBERT in all tasks.  Furthermore,
SBERT-WK offers the best performance in four of the ten tasks.  As
discussed in \cite{conneau2018you}, there is a tradeoff in shallow and
deep linguistic properties in a sentence.  That is, lower layer
representations carry more surface information while deep layer
representations represent more semantic meanings \cite{BERT_STRU}.
By merging information from various layers, SBERT-WK can take care of
these different aspects. 

The correlation between probing tasks and downstream tasks were studied
in \cite{conneau2018you}. They found that most downstream tasks only
correlates with a subset of the probing tasks. WC is positively
correlated with all downstream tasks. This indicates that the word
content (WC) in a sentence is the most important factor among all
linguistic properties. However, in our finding, although $p$-means
provides the best WC performance, it is not the best one in downstream
tasks. Based on the above discussion, we conclude that ``good
performance in WC alone does not guarantee satisfactory sentence
embedding and we should pay attention to the high level semantic meaning
as well".  Otherwise, averaging one-hot word embedding would give
perfect performance, which is however not true. 

The TREC dataset is shown to be highly correlated with a wide range of
probing tasks in \cite{conneau2018you}. SBERT-WK is better than SBERT in
all probing tasks and we expect it to yield excellent performance for
the TREC dataset. This is verified in Table \ref{exp:downstream_result}.
We see that SBERT-WK works well for the TREC dataset with substantial
improvement over the baseline SBERT model. 

SBERT is trained using the Siamese Network on top of the BERT model.  It
is interesting to point out that SBERT underperforms BERT in probing
tasks consistently. This could be attributed to that SBERT pays more
attention to the sentence-level information in its training objective.
It focuses more on sentence pair similarities. In contrast, the mask
language objective in BERT focuses more on word- or phrase-level and the
next sentence prediction objective captures the inter-sentence
information. Probing tasks are tested on the word-level information or
the inner structure of a sentence. They are not well captured by the
SBERT sentence embedding.  Yet, SBERT-WK can enhance SBERT significantly
through detailed analysis of each word representation. As a result,
SBERT-WK can obtain similar or even better results than BERT in probing
tasks. 

\subsection{Ablation and Sensitivity Study}

To verify the effectiveness of each module in the proposed SBERT-WK
model, we conduct the ablation study by adding one module at a time.
Also, the effect of two hyper parameters (the context window size and
the starting layer selection) is evaluated. The averaged results for
textual semantic similarity datasets, including STS12-STS16 and STSB,
are presented. 

\subsubsection{Ablation study of each module's contribution} 

We present the ablation study results in Table \ref{exp:ablation}.  It
shows that all three components (Alignment, Novelty, Token Importance)
improve the performance of the plain SBERT model. Adding the Alignment
weight and the Novelty weight alone provides performance improvement of
1.86 and 2.49, respectively. The Token Importance module can be
applied to the word representation of the last layer or the word
representation obtained by averaging all layer outputs. The
corresponding improvements are 0.55 and 2.2, respectively. Clearly,
all three modules contribute to the performance of SBERT-WK. The ultimate
performance gain can reach 3.56.

Table \ref{exp:attention_map} shows the attention heat maps of sentences from several different datasets. As we can see in the table, the word importance module indeed putting more focus on informative words.

\begin{table}[htb]
\centering
\caption{Comparison of different configurations to demonstrate the
effectiveness of each module of the proposed SBERT-WK method. The
averaged Pearson correlation coefficients ($\times 100 $) for STS12-STS16 and STSB
datasets are reported.}\label{exp:ablation}
                \resizebox{0.9\columnwidth}{!}{
                \begin{tabular}{ c | c } 
                    \hline
                    Model & Avg. STS results \\ \hline
                    SBERT baseline & 70.65\\ \hline
                    SBERT + Alignment ($w=0$) & 72.51\\ \hline
                    SBERT + Novelty ($w=1$) & 73.14 \\ \hline
                    SBERT + Token Importance (last layer) & 71.20 \\ \hline
                    SBERT + Token Importance (all layers) & 72.85 \\ \hline
                    SBERT-WK ($w=0.5$) & 74.21\\ 
                    \hline
                \end{tabular}}
\end{table}

\subsubsection{Sensitivity to window size and layer selection}

\begin{figure}[htb]
\centering
                \begin{subfigure}{0.49\columnwidth}
                    \centering
                    \includegraphics[width=\columnwidth]{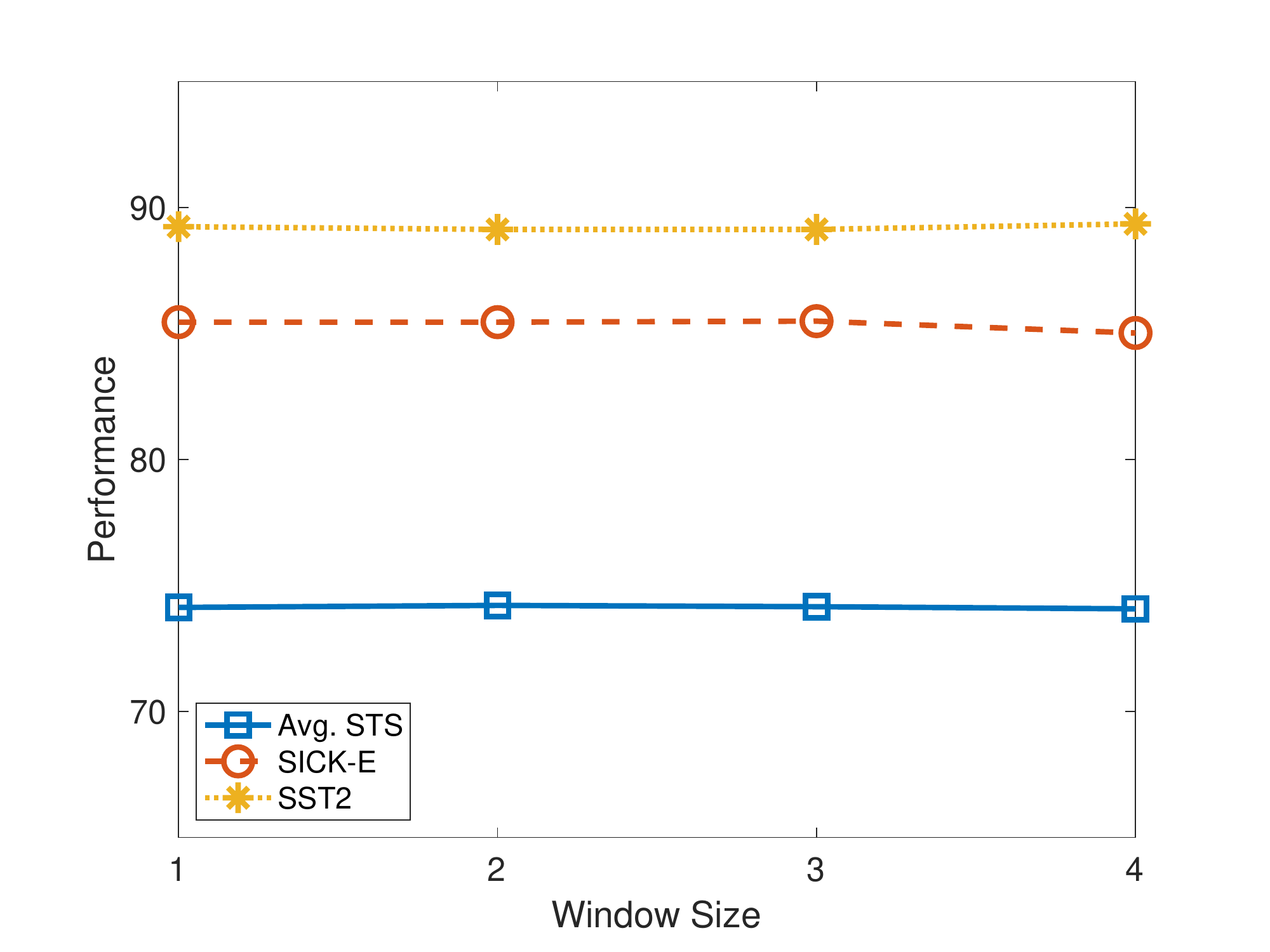}
                    \caption{}
                \end{subfigure} %
                \begin{subfigure}{0.49\columnwidth}
                    \centering
                    \includegraphics[width=\columnwidth]{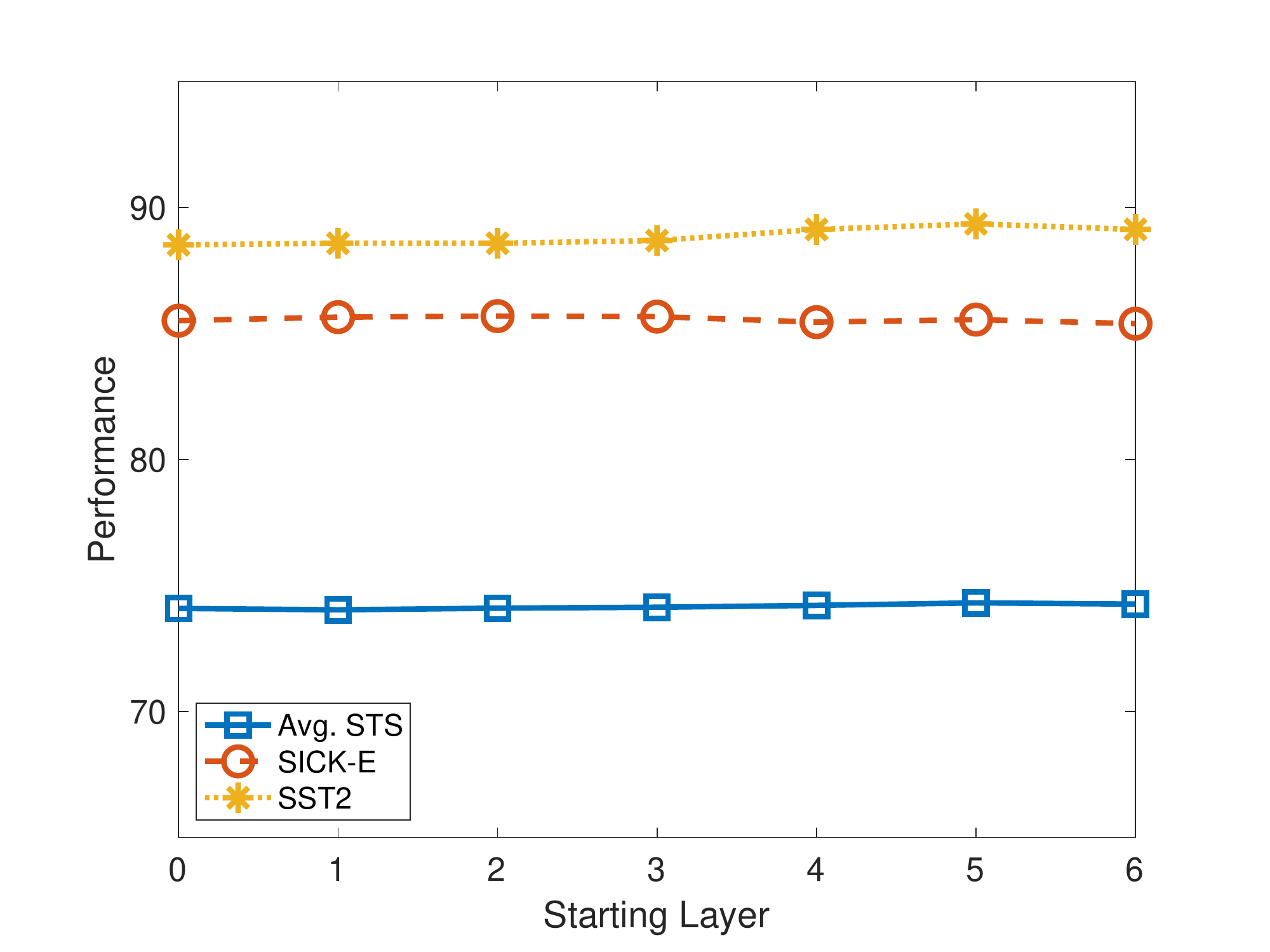}
                    \caption{}
                \end{subfigure}
\caption{Performance comparison with respect to (a) window size $m$ and
(b) starting layer $l_S$, where the performance for the STS datset is
the Pearson Correlation Coefficients ($\times100$) while the performance for the
SICK-E and the SST2 datasets is test accuracy. } \label{exp:sensitivity}
\end{figure}

We test the sensitivity of SBERT-WK to two hyper-parameters on STS,
SICK-E and SST2 datasets. The results are shown in Fig.
\ref{exp:sensitivity}. The window size $m$ is chosen to be 1, 2, 3 and
4.  There are at most 13 representations for a 12-layer transformer
network.  By setting window size to $m=4$, we can cover a wide range of
representations already. The performance versus the $m$ value is given
in Fig. \ref{exp:sensitivity} (a).  As mentioned before, since the first
several layers carry little contextualized information, it may not be
necessary to include representations in the first several layers. We
choose the starting layer $l_S$ to be from 0-6 in the sensitivity study.
The performance versus the $l_S$ value is given in Fig.
\ref{exp:sensitivity} (b).  We see from both figures that SBERT-WK is
robust to different values of $m$ and $l_S$. By considering the
performance and computational efficiency, we set window size $m=2$ as
the default value. For starting layer selection, the perform goes up a
little bit when the representations of first three layers are excluded.
This is especially true for the SST2 dataset.  Therefore, we set $l_S=4$
as the default value. These two default settings are used throughout 
all reported experiments in other subsections. 

\begin{table}[htb]
\centering
\caption{Inference time comparison of InferSent, BERT, XLNET, SBERT and
SBERT-WK. Data are collected from 5 trails.}\label{exp:speed}
                \resizebox{0.9\columnwidth}{!}{
                \begin{tabular}{ c | c | c } 
                    \hline
                    Model & CPU (ms) & GPU (ms) \\ \hline
                    InferSent (Conneau et al., 2017) & 53.07  & 15.23 \\ \hline
                    BERT (Devlin et al., 2018) & 86.89 & 15.27 \\ \hline
                    XLNET (Yang et al., 2018) & 112.49 & 20.98 \\ \hline
                    SBERT (Reimers et al., 2019) & 168.67 & 32.19 \\ \hline
                    SBERT-WK (SVD) & 179.27 & - \\ \hline
                    SBERT-WK (QR) & 177.26 & - \\ 
                    \hline
                \end{tabular}}
\end{table}

\begin{table*}[htb]
\centering
\caption{Attention heat map form SBERT-WK.}\label{exp:attention_map}
                \resizebox{1.5\columnwidth}{!}{
                \begin{tabular}{ c | c } 
                    \hline
                    Sentence Attention Map & Source \\ \hline
                    \colorbox{white!0}{
                    \colorbox{red!70.926517571885}{\strut authorities} \colorbox{red!24.28115015974441}{\strut in} \colorbox{red!46.96485623003196}{\strut ohio} \colorbox{red!29.07348242811502}{\strut ,} \colorbox{red!76.67731629392972}{\strut indiana} \colorbox{red!0.0}{\strut and} \colorbox{red!25.87859424920128}{\strut michigan} \colorbox{red!37.38019169329074}{\strut have} \colorbox{red!100.0}{\strut searched} \colorbox{red!84.02555910543133}{\strut for} \colorbox{red!43.76996805111822}{\strut the} \colorbox{red!29.392971246006383}{\strut bodies} 
                    } 
                    & STS  \\ \hline
                    \colorbox{white!0}{
                    \colorbox{red!64.89104116222761}{\strut anna} \colorbox{red!4.9636803874092}{\strut has} \colorbox{red!7.990314769975784}{\strut begun} \colorbox{red!0.0}{\strut to} \colorbox{red!22.154963680387407}{\strut rely} \colorbox{red!7.384987893462471}{\strut on} \colorbox{red!18.038740920096853}{\strut her} \colorbox{red!100.0}{\strut staler}\colorbox{red!7.384987893462471}{\strut 's} \colorbox{red!51.57384987893462}{\strut presence} 
                    } 
                    & SUBJ  \\ \hline
                    \colorbox{white!0}{
                    \colorbox{red!0.0}{\strut the} \colorbox{red!81.90789473684211}{\strut constitution} \colorbox{red!79.60526315789474}{\strut ality} \colorbox{red!26.97368421052631}{\strut of} \colorbox{red!100.0}{\strut outlawing} \colorbox{red!18.42105263157895}{\strut partial} \colorbox{red!51.315789473684205}{\strut birth} \colorbox{red!74.34210526315789}{\strut abortion} \colorbox{red!48.02631578947369}{\strut is} \colorbox{red!67.76315789473686}{\strut not} \colorbox{red!18.092105263157897}{\strut an} \colorbox{red!11.51315789473684}{\strut open} \colorbox{red!28.289473684210524}{\strut question} 
                    } 
                    & MPRC  \\ \hline
                    \colorbox{white!0}{
                    \colorbox{red!26.956521739130434}{\strut my} \colorbox{red!70.43478260869566}{\strut grandmother} \colorbox{red!19.206049149338366}{\strut barely} \colorbox{red!43.40264650283553}{\strut survived}
                    } 
                    & SubjNumber  \\ \hline
                \end{tabular}}
\end{table*}

\subsection{Inference Speed} \label{exp:computation}

We evaluate the inference speed against the STSB datasets. For fair
comparison, the batch size is set to 1.  All benchmarking methods are
run on CPU and GPU\footnote{Intel i7-5930K of 3.50GHz and Nvidia GeForce
GTX TITAN X are chosen to be the CPU and the GPU, respectively.}. Both
results are reported. On the other hand, we report CPU results of
SBERT-WK only.  All results are given in Table \ref{exp:speed}.  With
CPU, the total inference time of SBERT-WK (QR) is 8.59 ms (overhead)
plus 168.67ms (SBERT baseline).  As compared with the baseline BERT model,
the overhead is about 5\%.  SVD computation is slightly slower than QR
factorization.

\section{Conclusion and Future Work} \label{sec:conclusion}
    
In this work, we provided in-depth study of the evolving pattern of word
representations across layers in deep contextualized models.
Furthermore, we proposed a novel sentence embedding model, called
SBERT-WK, by dissecting deep contextualized models, leveraging the
diverse information learned in different layers for effective sentence
representations. SBERT-WK is efficient, and it demands no further
training.  Evaluation was conducted on a wide range of tasks to show the
effectiveness of SBERT-WK. 

Based on this foundation, we may explore several new research topics in
the future. Subspace analysis and geometric analysis are widely used in
distributional semantics. Post-processing of the static word embedding
spaces leads to furthermore improvements on downstream tasks
\cite{wang2018post,mu2017all}. Deep contextualized models have achieved
supreme performance in recent natural language processing tasks.  It
could be beneficial by incorporating subspace analysis in the deep
contextualized models to regulate the training or fine-tuning process.
This representation might yield even better results.  Another topic is
to understand deep contextualized neural models through subspace
analysis. Although deep contextualized models achieve significant
improvements, we still do not understand why these models are so
effective. Existing work that attempts to explain BERT and the
transformer architecture focuses on experimental evaluation.
Theoretical analysis of the subspaces learned by deep contextualized
models could be the key in revealing the myth. 

\ifCLASSOPTIONcaptionsoff
  \newpage
\fi

\bibliographystyle{IEEEtran}
\bibliography{IEEEabrv,mybibfile}

\vspace{-0.3in}

\begin{IEEEbiography}[{\includegraphics[width=1in,height=1.25in,clip,keepaspectratio]{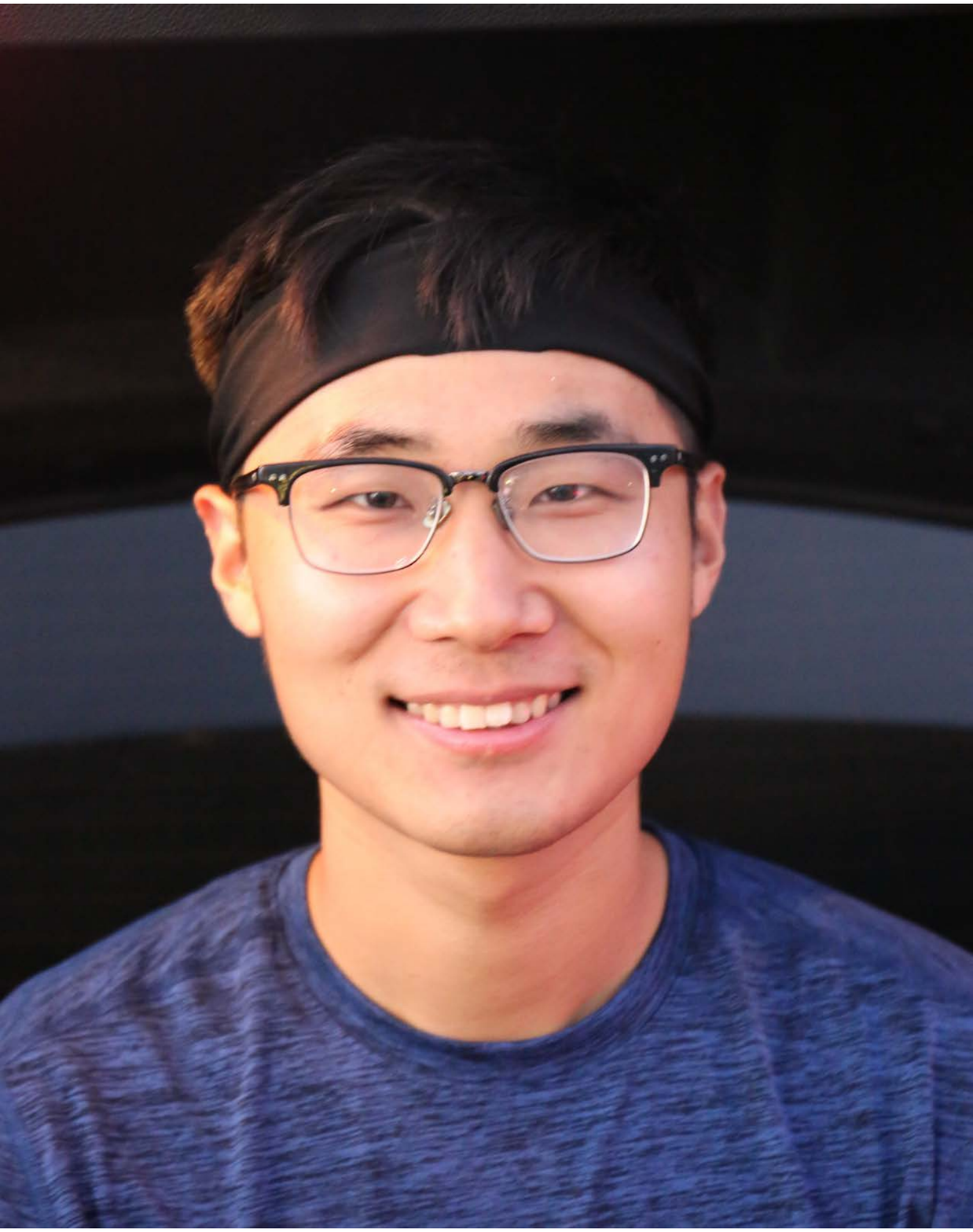}}]{Bin Wang}
Bin Wang received his Bachelor’s degree in Electronic Information
Engineering from University of Electronic Science and Technology of
China (UESTC), Chengdu, China in June, 2017. Since July 2017, He joined
Media Communication Lab (MCL) at University of Southern California (USC)
as a Ph.D. student, supervised by Prof. C.-C. Jay Kuo. His research
interests include natural language processing (NLP), image processing
and machine learning. 
\end{IEEEbiography}

\begin{IEEEbiography}[{\includegraphics[width=1in,height=1.25in,clip,keepaspectratio]{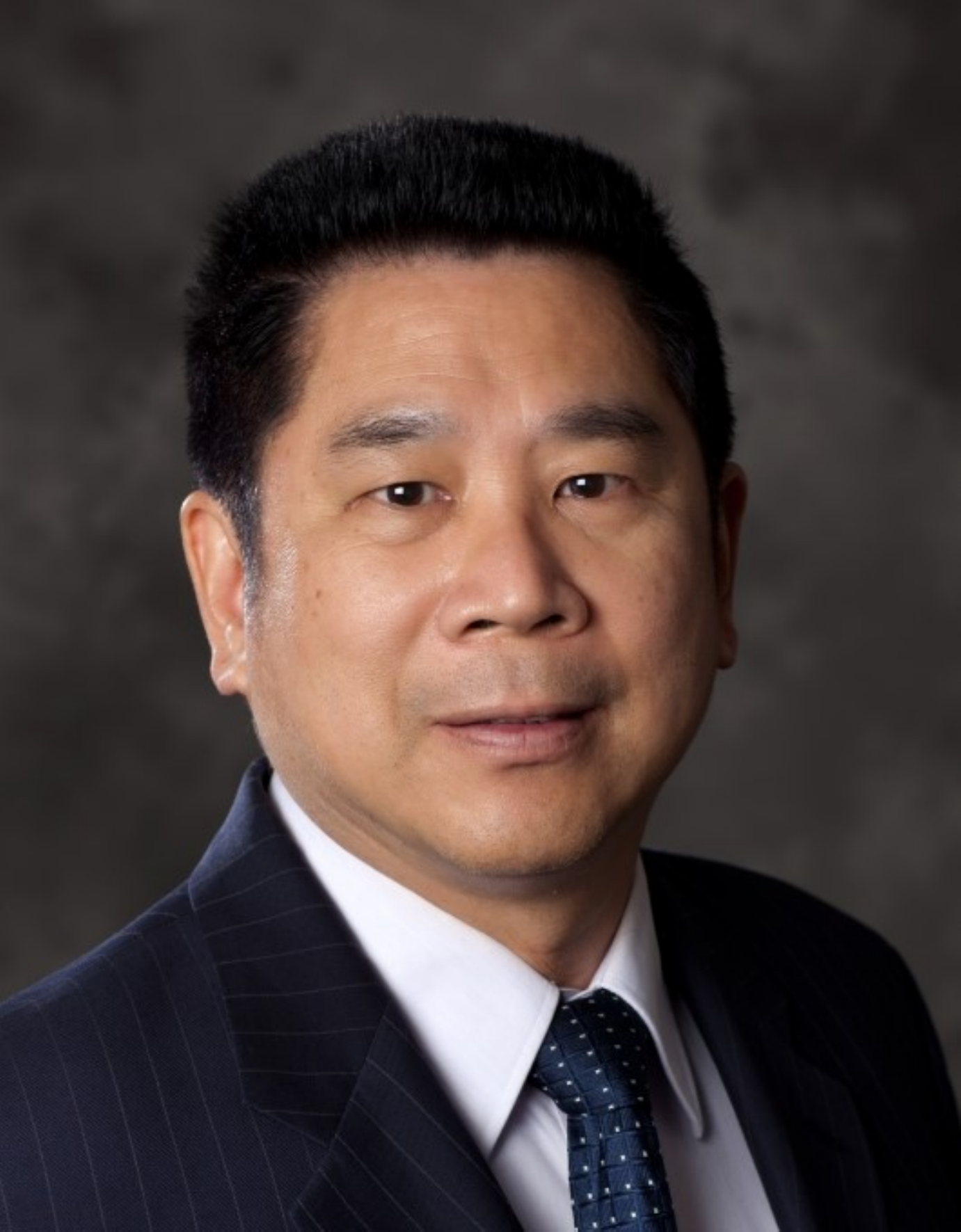}}]{C.-C. Jay Kuo}
C.-C. Jay Kuo (F’99) received the B.S. degree in electrical
engineering from the National Taiwan University, Taipei, Taiwan, in
1980, and the M.S. and Ph.D. degrees in electrical engineering from the
Massachusetts Institute of Technology, Cambridge, in 1985 and 1987,
respectively. He is currently the Director of the Multimedia
Communications Laboratory and a Distinguished Professor of electrical
engineering and computer science at the University of Southern
California, Los Angeles. His research interests include multimedia
computing and machine learning. Dr. Kuo is a Fellow of the American
Association for the Advancement of Science (AAAS) and The International
Society for Optical Engineers (SPIE).
\end{IEEEbiography}

\vfill
\end{document}